\documentclass[10pt,twocolumn,letterpaper]{article}

\usepackage{xcolor}
\usepackage{booktabs}
\usepackage{multirow}
\usepackage{listings} 

\usepackage{xcolor} 
\usepackage[cachedir=minted-cache]{minted}
\usepackage{textcomp} 
\usepackage[pagenumbers]{cvpr} 

\definecolor{cvprblue}{rgb}{0.21,0.49,0.74}
\usepackage[pagebackref,breaklinks,colorlinks,allcolors=cvprblue]{hyperref}

\def\paperID{5978} 
\def\confName{CVPR}
\def\confYear{2026}
\definecolor{shallowblue}{RGB}{227, 240, 249}
\definecolor{deepblue}{RGB}{192, 216, 240}
\title{OmniGround: A Comprehensive Spatio-Temporal Grounding Benchmark for Real-World Complex Scenarios}

\begin{document}
\author{
Hong Gao$^{1,2}$\thanks{Equal Contribution.}\quad
Jingyu Wu$^2$\footnotemark[1]\quad
Xiangkai Xu$^2$\quad
Kangni Xie$^2$\quad
Yunchen Zhang$^2$\quad \\
Bin Zhong$^2$\quad
Xurui Gao$^1$\quad
Min-Ling Zhang$^1$\thanks{Corresponding Author}\\
$^{1}$SouthEast University\quad $^2$ZTE Corporation}

\twocolumn[{
\renewcommand\twocolumn[1][]{#1}

\maketitle

\begin{center}
    \centering
    \label{fig:base_case}
    \captionsetup{type=figure}
    \includegraphics[width=\textwidth]{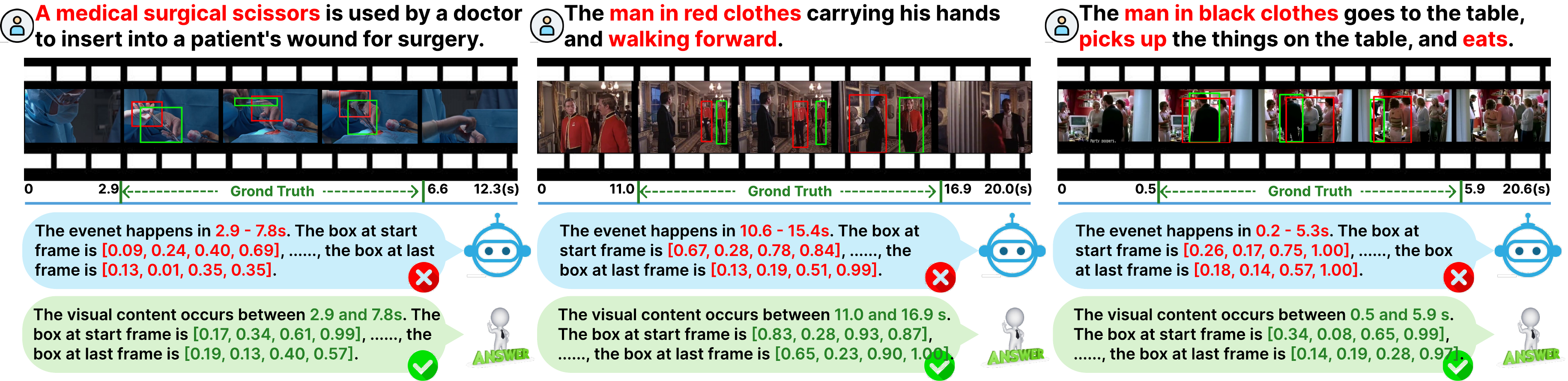}
    \captionof{figure}
    {Some examples of STVG MLLMs facing real-world complex scenarios (left: uncommon objects, middle: multiple similar target objects, right: queries with deep syntactic complexity). Content in \textcolor{green}{green} and \textcolor{red}{red} represents correct and wrong answers, respectively.}
        \label{fig:teaser}
\end{center}%
}]

\begin{abstract}
Spatio-Temporal Video Grounding (STVG) aims to localize target objects in videos based on natural language descriptions. 
Despite recent advances in Multimodal Large Language Models, a significant gap remains between current models and real-world demands involving diverse objects and complex queries. 
We attribute this to limited benchmark scope, causing models to exhibit category bias, oversimplified reasoning, and poor linguistic robustness.
To address these limitations, we introduce OmniGround, a comprehensive benchmark with 3,475 videos spanning 81 categories and complex real-world queries. We propose the Forward-Backward-Refinement annotation pipeline that combines multi-directional tracking with intelligent error correction for high-quality labels.
We further introduce DeepSTG, a systematic evaluation framework quantifying dataset quality across four complementary dimensions beyond superficial statistics.
Evaluations reveal performance average drop of 10.4\% on complex real-world scenes, particularly with small/occluded objects and intricate spatial relations. Motivated by these, we propose PG-TAF, a training-free two-stage framework decomposing STVG into high-level temporal grounding and fine-grained spatio-temporal propagation. Experiments demonstrate PG-TAF achieves 25.6\% and 35.6\% improvements in m\_tIoU and m\_vIoU on OmniGround with consistent gains across four benchmarks. 
\end{abstract}    
\section{Introduction}
\label{sec:intro}

Spatio-Temporal Video Grounding (STVG) aims to localize target objects in both temporal and spatial domains within a video based on natural language descriptions~\cite{zhang2020does}.
It finds broad applications in autonomous driving~\cite{vishal2024eyes,ahmad2025videomolmo}, video retrieval~\cite{gu2024context,yang2025multio}, intelligent surveillance~\cite{zhang2020does}, and human-computer interaction~\cite{zhou2023language}.
With the rapid development of Multimodal Large Language Models (MLLMs)~\cite{wang2025capabilities,comanici2025gemini,yang2025qwen3,dubey2024llama}, the task has greatly benefited from MLLMs' enhanced cross-modal semantic understanding and structured spatial reasoning, substantially outperforming traditional lightweight approaches~\cite{cai2025has,huang2024vtimellm,li2024groundinggpt,wang2023visionllm}.
\begin{figure}[htbp]  
\centering  
\includegraphics[width=0.5\textwidth]{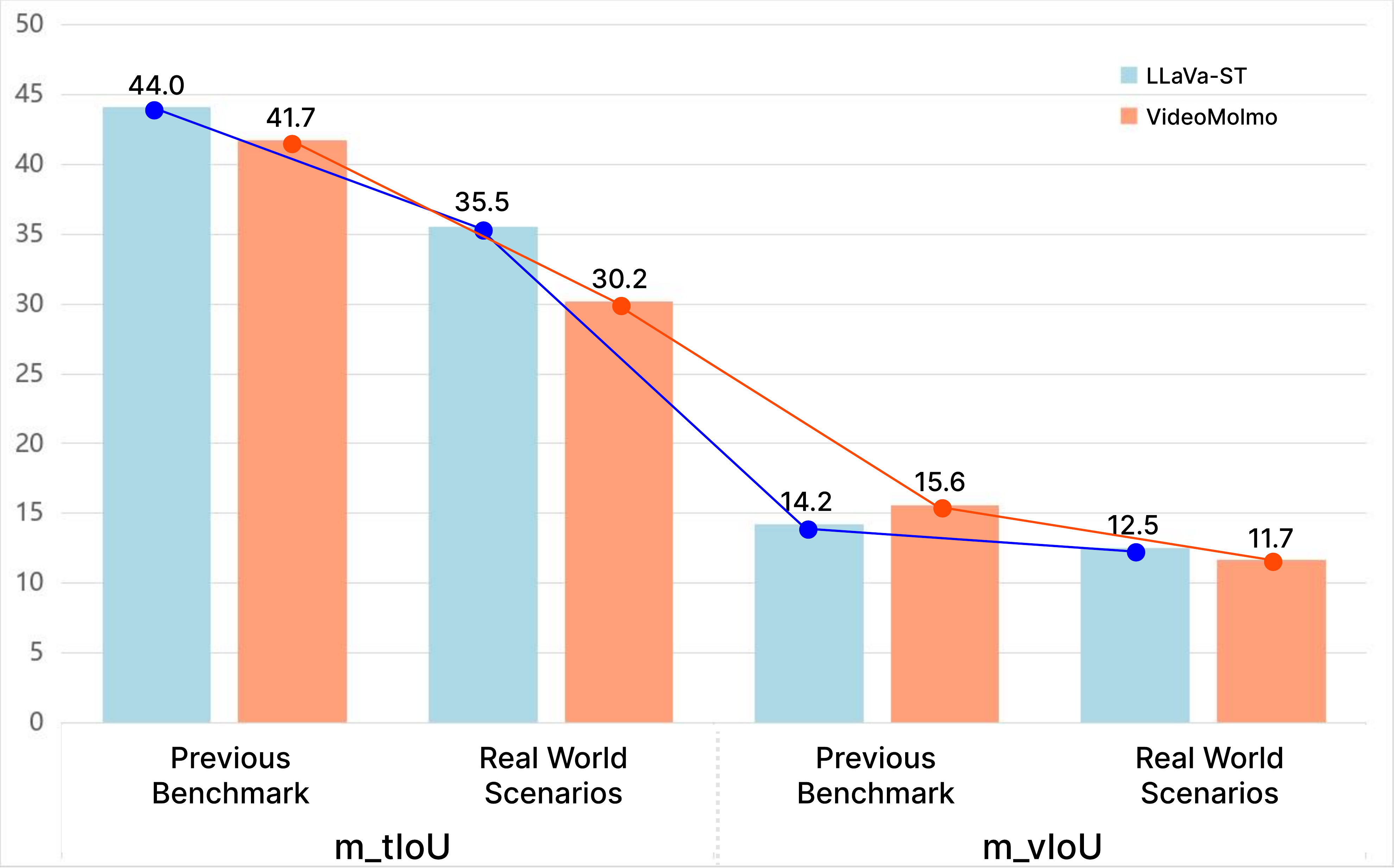}  
\caption{Performance of STVG MLLMs on complex real-world inputs. Inconsistent trends indicate current datasets are insufficient for comprehensive real-world evaluation.}
\label{fig:badcase_chart}  
\end{figure}  

Despite these advances, there remains a significant gap between current STVG-MLLMs and real-world demands~\cite{wang2025spacevllm,ahmad2025videomolmo,li2025llava}. 
As illustrated in~\cref{fig:teaser}, existing models often struggle with complex real-world scenarios: failing to localize some uncommon objects (e.g., ``kite", ``scissors"); unable to distinguish multiple similar targets (``the right man in red clothes among three people"); and misinterpreting queries with nested spatial relations.

We attribute these limitations to the narrow scope of current benchmarks. 
Datasets like HC-STVG~\cite{tang2021human} focus predominantly on a single category (person), while others like VidSTG~\cite{zhang2020does,shang2019annotating} involve relatively simple scenarios with limited object diversity, which hinders the practical deployment of STVG-MLLMs in complex real-world scenarios.

To address these limitations, we introduce \textbf{OmniGround}, a comprehensive benchmark designed to advance STVG-MLLMs toward real-world applicability. 
OmniGround comprises 3,475 videos (average 18.2s) spanning 81 balanced object categories — twice as many as VidSTG~\cite{zhang2020does} and six times more than ST-Align~\cite{li2025llava}.
Each video is densely annotated with spatio-temporal tubes at native FPS, paired with linguistically complex descriptions (sentence-level NEI: 0.918 vs. 0.659 in VidSTG) that include rich spatial relations and diverse action predicates.

To ensure annotation quality and efficiency, we propose the \textbf{Forward-Backward-Refinement (FBR)} pipeline, which uses multi-point bi-directional tracking with fusion-based voting to generate accurate spatio-temporal tubes.
Our FBR pipeline reduces annotation time consumption while significantly enhancing continuity and accuracy, and is particularly robust to partial occlusions (\cref{fig:anno_pipeline}).

Furthermore, to systematically evaluate dataset quality beyond superficial statistics (e.g., video duration and clip counts), we introduce \textbf{DeepSTG}, an evaluation framework that quantifies high-dimensional metrics such as category diversity, foreground complexity, linguistic balance, and cross-modal alignment.
A comprehensive analysis with DeepSTG demonstrates that OmniGround significantly outperforms existing benchmarks across all metrics (\cref{tab:comparison_DeepSTG}), thereby establishing a new standard for STVG research.

Based on OmniGround, we conduct extensive evaluations on multiple state-of-the-art models to discover their current performance limitations.
Models exhibit performance average drop of {10.4}\% in m\_vIoU compared to simpler benchmarks (\cref{fig:badcase_chart}), particularly in: \textit{(i)} categories bias ({8.42}\% drop), \textit{(ii)} oversimplified reasoning ({13.1}\% drop), and \textit{(iii)} poor robustness to linguistic complexity ({9.42}\% drop).
The results reveal that existing methods struggle with samples exhibiting high foreground complexity and deep semantic structure, particularly when the object is small or heavily occluded, or when spatial relations are intricate.

Motivated by these observations, we propose \textbf{PG-TAF}, a training-free two-stage framework that decouples STVG into coarse temporal grounding (via MLLM's holistic reasoning) and fine-grained spatial propagation (via specialized tracking), effectively addressing the reasoning-localization trade-off. 
Experiments across four benchmarks demonstrate consistent improvements, with 25.6\% and 35.6\% gains in m\_tIoU and m\_vIoU on OmniGround.
We hope our work can inspire ideas for improving STVG performance under complex real-world scenarios.

In summary, the contributions of our work are as follows:

\begin{itemize}[leftmargin=*]

    \item We construct \textbf{OmniGround}, a freely available STVG benchmark comprising 3,475 videos with 81 balanced categories and complex real-world queries. We further introduce the \textbf{FBR pipeline} for efficient annotation.

    \item We propose \textbf{DeepSTG}, a systematic framework for quantifying STVG dataset quality, revealing limitations in category diversity, complex reasoning, and language robustness. Evaluations on OmniGround further show that current models struggle in these complex scenarios due to these constraints in existing datasets.

    \item We propose \textbf{PG-TAF}, a training-free two-stage framework baseline designed to address the challenges exposed by OmniGround.
    Experiments on four benchmarks compared with several models indicate improvements in our method and a promising direction for future research. 
    
\end{itemize}

\section{Related Work}
\label{sec:rw}

\subsection{Spatio-Temporal Video Grounding Datasets}
\label{sec:2.1}
STVG research relies on comprehensive datasets to improve model performance. HC-STVG~\cite{tang2021human} is a pioneering large-scale benchmark that localizes spatio-temporal tubes of target persons in untrimmed videos based on textual descriptions. VidSTG~\cite{zhang2020does} provides an untrimmed video benchmark emphasizing multi-form sentence grounding. Li et al.\cite{li2025llava} propose ST-Align, using GPT-4-turbo\cite{achiam2023gpt} to enhance VidSTG's textual annotations. Ahmad et al.\cite{ahmad2025videomolmo} introduce VPos-Bench with 72k video-caption pairs and 100k annotated object points across five real-world scenarios. EgoMask\cite{liang2025fine} explores fine-grained object tracking in egocentric videos based on Ego4D~\cite{grauman2022ego4d}.

Despite these efforts, existing benchmarks lack comprehensive diagnostic evaluation beyond basic statistics.
Current assessments of STVG benchmarks primarily rely on basic statistical metrics, such as object type, domain, and number~\cite{su2021stvgbert}. 
While HC-STVG~\cite{tang2021human} introduces concepts like Ambiguity (BA, TA) and High-level Compositionality (HC), these metrics still can not fully evaluate the benchmark's complexity and generalization demands~\cite{WANG2023efficient}.
Specifically, the ambiguity metrics use binary flags or broad categories that fail to capture the true level of difficulty in the data.
HC is often based on simple feature counts instead of balanced syntactic challenges, leading to uneven data distributions that hurt model performance.

Therefore, we introduce \textbf{OmniGround}, a comprehensive STVG benchmark designed to address the lack of diversity and complexity in existing datasets.
Furthermore, to assess dataset quality beyond superficial statistics, we propose \textbf{DeepSTG}, an evaluation framework that enabling objective comparison and rigorous analysis of benchmarks.

\subsection{Spatio-Temporal Video Grounding Via MLLM}
With the significant progress of MLLMs in video understanding, more research focuses on using MLLMs to effectively address the STVG task~\cite{li2024groundinggpt,wang2025spacevllm,ahmad2025videomolmo,li2025llava}. 
Existing methods can be categorized into two main perspectives.

The first perspective decomposes STVG into multi-stage processes. Methods like VTimeLLM~\cite{huang2024vtimellm} and TRACE~\cite{guo2024trace} focus on temporal localization through boundary-sensitive training or causal event modeling.
VideoMolmo~\cite{ahmad2025videomolmo} focuses on accurately grounding the query in time. 
Groma~\cite{ma2024groma} designs a localized visual tokenization mechanism for fine-grained region captioning, and GroundingGPT~\cite{li2024groundinggpt} uses MLLM to pinpoint the target object's location spatially.

The second perspective is to let MLLMs simultaneously capture both temporal and spatial information.
Recently, LLaVA-ST~\cite{li2025llava} proposes language-aligned positional embedding and a spatial-temporal packer to improve spatio-temporal multimodal understanding. 
SpaceVLLM~\cite{wang2025spacevllm} adopts spatio-temporal aware queries and a query-guided space decoder to enhance the MLLMs' performance.

Despite recent advances, existing methods struggle with multi-actor videos, complex foregrounds, and linguistically difficult queries~\cite{wang2025spacevllm,ahmad2025videomolmo}. 
Because current STVG datasets focus on simple, clean video segments without real-world complexity~\cite{tang2021human,zhang2020does,li2025llava}. 
Therefore, we propose OmniGround to comprehensively evaluate the models' performance under complex conditions.

\section{OmniGround Benchmark}

\begin{figure}[htbp]  
\centering  
\includegraphics[width=0.5\textwidth]{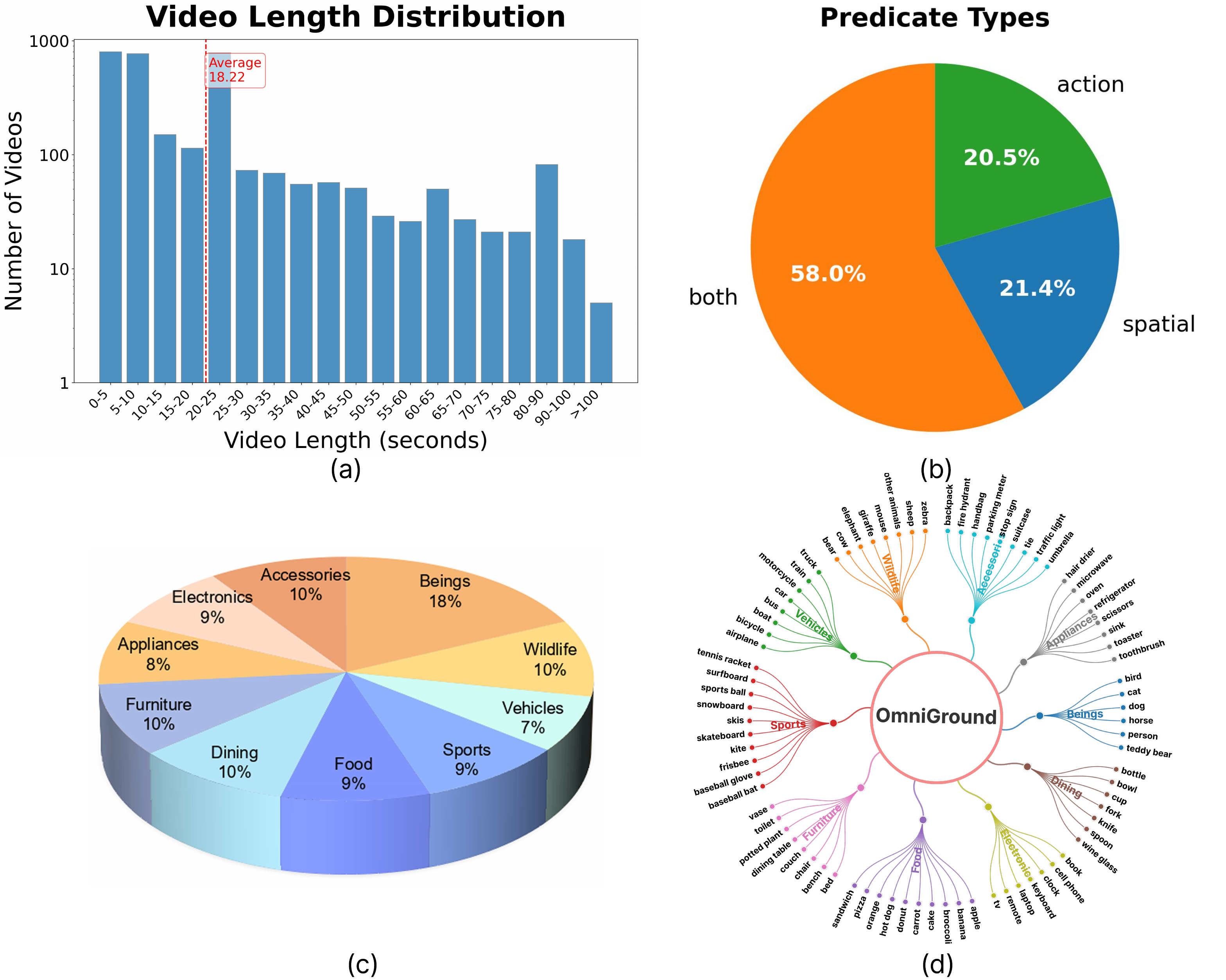}  
\caption{Statistics of our OmniGround benchmark: (a) Video durations range from short clips to 140 seconds. (b) Predicates combine spatial and action elements, reflecting complex object interactions and motion. (c) Balanced category distribution. (d) 81 object categories.}
\label{fig:base_ana}  
\end{figure}

In this section, we introduce the construction pipeline of our OmniGround, which consists of four key stages:1) data acquisition, 2) data manipulation and filtering, 3) high-quality annotation via the Forward-Backward-Refinement pipeline, and 4) dataset augmentation and validation to enhance coverage. 
\cref{fig:base_ana} shows the overall statistics of OmniGround.

\subsection{Data Acquisition}
\label{sec:3.1}
We use a two-source strategy to ensure diversity, real-world relevance, copyright compliance, and annotation feasibility:

\textbf{Internet Videos.} We collect high-quality, copyright-free videos from video-sharing platforms like Pexels, covering diverse real-world scenarios. 
To ensure category balance, we select at least 5 videos each from 81 categories. 
We prioritize videos that exhibit: \textit{(i)} natural lighting and camera motion, \textit{(ii)} multiple interacting objects, and \textit{(iii)} complex backgrounds typical of uncontrolled environments.

\textbf{RVOS Datasets Augmentation.} 
To improve coverage of rare object classes (e.g., “keyboard”), we slightly supplement with RVOS videos~\cite{seo2020urvos,ding2023mevis}, which are processed (\cref{sec:3.4}) to fit STVG tasks.

\subsection{Data Manipulation}

Following data acquisition, we establish two temporal constraints to ensure data quality and task validity:
(1) Video Duration $\geq$ 3s: Videos shorter than 3s lack sufficient temporal context for STVG queries, making temporal grounding trivial.
(2) Event Duration $\geq$ 1s: Target events lasting less than 1s are too fleeting for reliable annotation and evaluation, as they approach the granularity of single frames.

After filtering, OmniGround comprises 3,475 videos with an average duration of 18.2 seconds (ranging from 3s to 140s). 
Each video contains an independent annotated spatio-temporal event, with caption averaging 16.2 words. 
The benchmark exhibits high diversity across object categories (81 balanced classes) and rich predicate structures.

\begin{figure*}[htbp]  
\centering  
\includegraphics[width=0.95\textwidth]{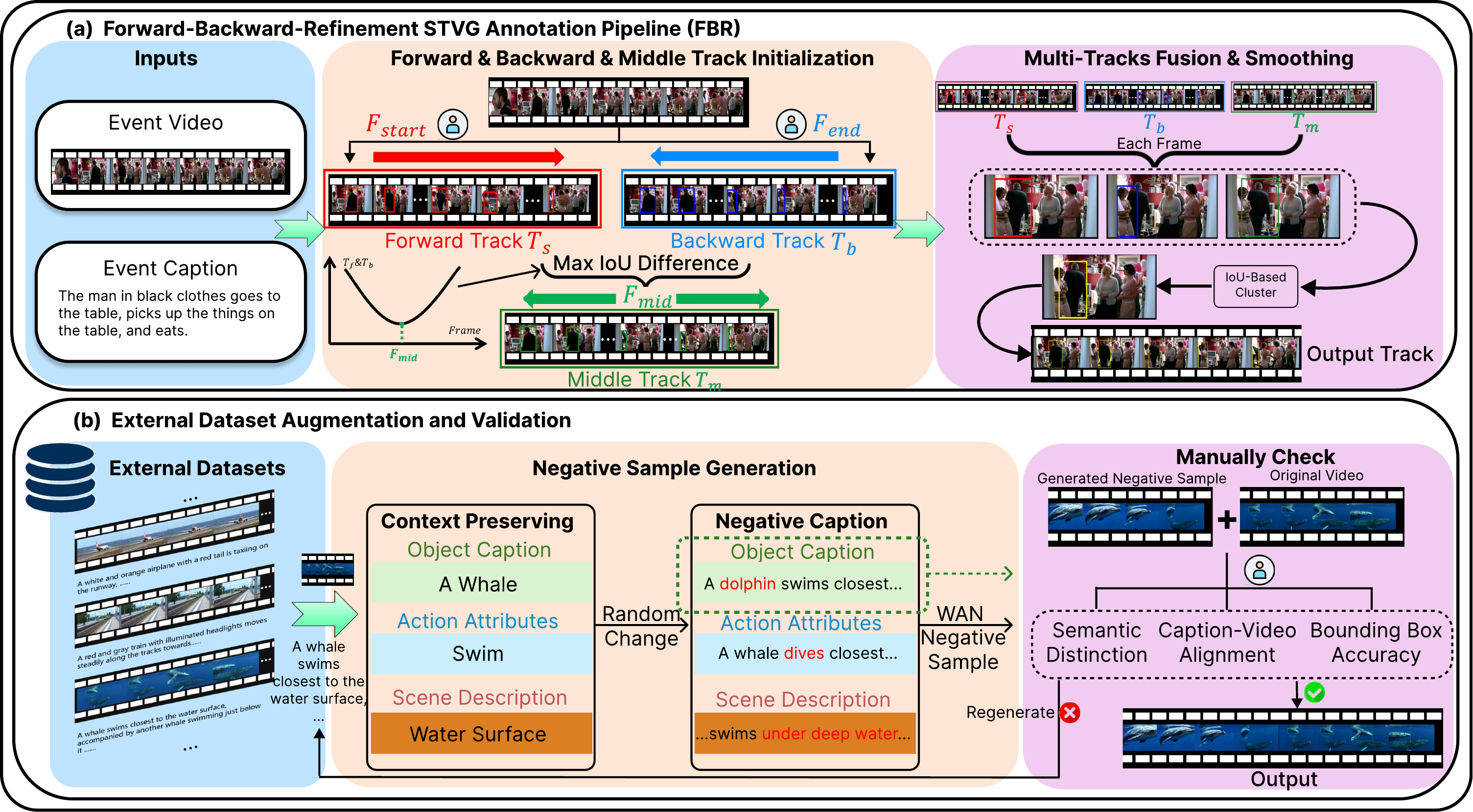}  
\caption{(a) Overview of our Forward-Backward-Refinement pipeline. (b) An illustration of external data augmentation and validation.}  
\label{fig:anno_pipeline}  
\end{figure*}

\subsection{Forward-Backward-Refinement Pipeline}
\label{sec:3.3}
Accurate spatio-temporal tube annotation faces three challenges: high labor cost from frame-by-frame manual labeling; tracking drift due to occlusions and motion blur; and annotation discontinuity from accumulated errors. 
To address these challenges, we propose the Forward-Backward-Refinement (FBR) pipeline (\cref{fig:anno_pipeline} (a)), which combines multi-directional tracking with intelligent error correction:

\begin{itemize}
    \item \textbf{Step 1: Multi-Point Initialization.} Instead of relying on a single starting point, annotators first manually label bounding boxes at two key frames: $F_{start}$ and $F_{end}$. 
    We then initialize bi-directional tracking from $F_{start}$ and $F_{end}$ using DAM4SAM tracker~\cite{videnovic2025distractor}, generating forward track $T_f$ and backward track $T_b$.

    \item \textbf{Step 2: Adaptive Refinement.} The pipeline automatically identifies the frame $F_{mid}$ with maximum IoU discrepancy between $T_f$ and $T_b$, indicating potential tracking failure. 
    The annotator then provides a corrective bbox at $F_{mid}$, from which we generate middle track $T_m$.

    \item \textbf{Step 3: Fusion and Smoothing.}
    The three tracks are fused via IoU-based voting: bboxes within each frame are clustered (IoU $\geq$ 0.5), and the main cluster's average serves as the fused result. 
    For short gaps where the number of continuously lost frames is $\leq$ 3, the pipeline recovers them by interpolating between continuous segments using Kalman filtering~\cite{simon2001kalman}, allowing the track to persist through brief occlusions.

\end{itemize}

Compared to single-direction tracking, FBR improves IoU consistency by 16.8\% (measured on 100 validation samples). 
The pipeline is particularly robust to partial occlusions, maintaining $\geq$ 0.8 IoU even when the target is 50\% occluded.
Additionally, we adopt the real-time quality monitoring protocol from~\cite{lotfian2017building,burmania2015increasing}, automatically pausing annotators when their consistency drops below threshold.

\subsection{External Dataset Augmentation and Validation}
\label{sec:3.4}
Videos sourced from RVOS datasets often lack the temporal boundaries and negative samples essential for STVG.
This is because RVOS videos typically feature annotations that span the entire duration. 
To address this, we introduce an automated negative sample generation pipeline (\cref{fig:anno_pipeline} (b)).
For each video with full-duration annotations, we:
\begin{enumerate}[leftmargin=*]
    \item \textbf{Semantic aspect selection.} To create reasonable distractors while ensuring context consistency, we randomly select \textit{one} of three semantic aspects to modify: \textit{i) object} (e.g., ``soccer ball" $\rightarrow$ ``basketball"), \textit{ii) action} (e.g., ``kicking" $\rightarrow$ ``holding"), or \textit{iii) scene} (e.g., ``on a beach" $\rightarrow$ ``in a park"), while preserving the other two aspects.
    
    \item \textbf{Counterfactual caption generation.} Using LLMs (Qwen3~\cite{yang2025qwen3} or DeepSeek~\cite{liu2024deepseek}), we generate modified captions based on the selected aspect. 
    For example, ``person kicking a soccer ball on a beach" may become ``person kicking a basketball on a beach" (object change) while keeping action and scene unchanged.
    
    \item \textbf{Negative video synthesis.} We take the counterfactual caption as the input of video generation model WAN~\cite{wan2025wan} to synthesize negative video clips that visually match the original context but exhibit the modified semantic aspect. 
    These generated clips are temporally concatenated to the original video—as a prefix or suffix—creating plausible but incorrect temporal segments.
    
    \item \textbf{Manual validation.} All generated samples undergo manual validation to ensure:
    \textit{(i)} clear semantic distinction from the ground truth,
    \textit{(ii)} caption-video alignment verification (using Keye-VL~\cite{team2025kwai} for ambiguous cases), and
    \textit{(iii)} bounding box accuracy (refined via FBR if needed).
    Samples that fail validation are regenerated with more significant modifications.
\end{enumerate}

\section{DeepSTG Evaluation System}

In this section, we introduce \textbf{DeepSTG}, a systematic evaluation framework that quantifies dataset quality across four complementary dimension.
 We apply DeepSTG to comprehensively evaluate seven benchmarks, demonstrating OmniGround significantly outperforms across all metrics.

\begin{table*}[h]
\small
    \centering
    \caption{Results of dataset quality assessment using the DeepSTG evaluation system. `\#' indicates the number.}
    \label{tab:comparison_DeepSTG}
    \begin{tabular}{lccccccc}
        \toprule
        \multirow{2}{*}{\textbf{Benchmark}} & \multirow{2}{*}{\textbf{\#Category}} & \multicolumn{3}{c}{\textbf{NEI}} &\multirow{2}{*}{\textbf{FCI}} & \multirow{2}{*}{\textbf{VSBI}} & \multirow{2}{*}{\textbf{CMA Score}} \\
        \cmidrule(lr){3-5} 
        & & \textbf{Category} & \textbf{Video Duration} & \textbf{Length of Sentence} & & & \\
        \midrule 
        MeVis & 14 &\colorbox{shallowblue}{0.952} & 0.421& 0.635& 0.728 & \colorbox{shallowblue}{0.753}& 0.624\\
        Ref-YT-VOS & 38 & 0.900& - & \colorbox{shallowblue}{0.859}& 0.719 & 0.410&0.757 \\
        VastTrack & \colorbox{shallowblue}{55} & 0.724 & 0.520  & 0.773 &0.736 & 0.679 &\colorbox{shallowblue}{0.779}\\
        HC-STVG & 1 & -& -& 0.778& \colorbox{shallowblue}{0.754} &0.431 &0.678 \\
        VIDSTG & 40 &0.569 &\colorbox{shallowblue}{0.780} & 0.659& 0.720 & 0.699& 0.701\\
        ST-Align & 14 & 0.628& 0.731&0.757 & 0.718 &0.466 &0.644 \\
        \midrule
        OmniGround (ours) & \colorbox{deepblue}{81} & \colorbox{deepblue}{0.992}& \colorbox{deepblue}{0.807}& \colorbox{deepblue}{0.918}& \colorbox{deepblue}{0.770}  & \colorbox{deepblue}{0.897} & \colorbox{deepblue}{0.813}\\
        \bottomrule 
    \end{tabular}
\end{table*}

\subsection{DeepSTG Evaluation System Design}
Our evaluation system is guided by three design principles:

\textbf{(1) Multi-Faceted Assessment:} No single metric can capture dataset quality. 
DeepSTG combines multiple dimensions, including annotation fidelity (CMA), visual challenge (FCI), linguistic diversity (VSBI), and distributional balance (NEI), to provide holistic evaluation.

\textbf{(2) Task-Aware Design:} Each of four selected metrics directly evaluate a critical part of the STVG task: CMA ensures the fundamental correctness of annotations, which is essential for effective supervised learning. 
FCI quantifies the difficulty of spatial discrimination, directly impacting the model's localization accuracy. 
VSBI measures the balance of linguistic cues (temporal/spatial), ensuring a balanced requirement for temporal and spatial reasoning. 
NEI guards against dataset bias, which is critical for ensuring the model's generalization ability across different scenarios.

\textbf{(3) Automation and Objectivity:} DeepSTG uses automated metrics based on authoritative measures to ensure reproducibility and eliminate subjective bias.

\subsubsection{Cross-Modal Semantic Alignment Score}
Ensuring consistency between natural language descriptions and their spatio-temporal tubes is crucial for verifying annotation quality. Standard manual verification lacks both scalability and objectivity. To address this, we introduce the \textbf{Cross-Modal Semantic Alignment} (CMA) Score, which leverages MLLM reasoning to automatically assess the semantic coherence among the video, caption, and bboxes.

Specifically, for a given video-caption-tube(bbox) triplet $(V, Q, B)$, we sample $N$ key frames $\mathcal{F} = \{f_1, f_2, \ldots, f_N\}$ within the temporal segment $[T_{start}, T_{end}]$ defined by tube $B$. 
For each frame $f_i$ and its bbox $b_i$, we prompt GPT-4o~\cite{achiam2023gpt} to evaluate three semantic aspects:

\textbf{(1) Object Presence Score ($\text{S}_{obj}$):} Does the target object mentioned in $Q$ exist within $b_i$?

\textbf{(2) Action Accuracy Score ($\text{S}_{act}$):} Is the action/state described in $Q$ accurately performed by the object in $b_i$?

 \textbf{(3) Context Consistency Score ($\text{S}_{ctx}$):} Are the spatial/temporal relations described in $Q$ consistent with the visual context in $f_i$ and video $V$?

Each aspect receives a score in $[0, 1]$. 
The CMA score averages across all frames and aspects, and higher scores indicate better annotation quality and semantic integrity:

\begin{equation} 
\text{CMA} = \frac{1}{N} \sum_{i=1}^{N} \left[ \frac{\text{S}_{obj}(f_i) + \text{S}_{act}(f_i) + \text{S}_{ctx}(f_i)}{3} \right]
\end{equation}

\subsubsection{Foreground Complexity Index}
STVG models must distinguish targets from visually similar distractors, especially when multiple same-class instances coexist (e.g., the left red car among three red cars). To measure visual clutter, we propose the \textbf{Foreground Complexity Index} (FCI), which quantifies intra-class visual similarity.

Given video $V$, we sample 1 frame per second and detect all foreground objects using YOLOv11~\cite{khanam2024yolov11}, obtaining object set $\mathcal{O} = \{o_i\}_{i=1}^{N}$ with feature vectors $\mathcal{V} = \{v_i\}_{i=1}^{N}$. 
For each category $C$, we compute intra-class mean similarity:
\begin{equation}
S_{\text{in}}(C) = \frac{1}{|\mathcal{O}_C|^2 - |\mathcal{O}_C|} \sum_{o_i, o_j \in \mathcal{O}_C, i \neq j} \text{CosSim}(v_i, v_j)
\end{equation}
where $\mathcal{O}_C$ denotes objects of category $C$ (computed only if $|\mathcal{O}_C| \geq 2$). 
The FCI averages across all categories:
\begin{equation}
\text{FCI} = \frac{1}{|\mathcal{C}|} \sum_{C \in \mathcal{C}} S_{\text{in}}(C), \text{FCI}\in [0, 1]
\end{equation}
where higher values indicate objects within the same category are visually similar (harder to distinguish), while lower values suggest diverse appearances (easier grounding).

\subsubsection{Verb-Spatial Balance Index}
STVG queries require action (``running") and spatial (``in front of") reasoning. 
Imbalanced datasets biased toward one type hinder model generalization. 
We introduce \textbf{Verb-Spatial Balance Index (VSBI)} to measure linguistic diversity.
We categorize each word in captions into three types:

\begin{itemize}[leftmargin=*]
    \item \textbf{Action Cues ($\mathcal{A}$):} Actions (``run") and states (``stand")
    \item \textbf{Spatial Cues ($\mathcal{S}$):} Directional (``towards, away") and locational (``behind, next to")
    \item \textbf{Mixed Cues ($\mathcal{M}$):} Combining both (``climbing up")
\end{itemize}

Let $P_{\text{actual}} = [P_\mathcal{A}, P_\mathcal{S}, P_\mathcal{M}]$ denote the normalized frequency distribution, where $P_\mathcal{A} + P_\mathcal{S} + P_\mathcal{M} = 1$. 
The ideal balanced distribution is $P_{\text{ideal}} = [1/4, 1/4, 1/2]$. VSBI $\in [0,1]$ measures distance from this ideal:
\begin{equation}
\text{VSBI} = 1 - \frac{\|P_{\text{actual}} - P_{\text{ideal}}\|_2}{\sqrt{7/8}}
\end{equation}
where $\sqrt{7/8}$ is the maximum possible Euclidean distance.

\subsubsection{Normalized Entropy Index}
To quantify the distributional uniformity across object category, video duration, and text description length, we introduce the Normalized Entropy Index (NEI).
The NEI evaluates the balance of data, where a higher score indicates less biased distribution across the dataset:
\begin{equation}
\text{NEI} = \frac{H}{\log N},H = -\sum_{i=1}^{N} p_i \log p_i
\end{equation}
where $H,\log N, p_i,$ represents the Shannon Entropy~\cite{lin2002divergence}, Maximum Entropy~\cite{wu2012maximum}, and relative frequency of an instance falling into a specific range (e.g., a specific category).

\subsection{STVG Dataset Analyze and Comparison}
We apply DeepSTG to evaluate seven benchmarks: four STVG benchmarks (HC-STVGv2~\cite{tang2021human}, VidSTG~\cite{zhang2020does}, ST-Align~\cite{li2025llava}, OmniGround) and three RVOS benchmarks (MeVis~\cite{ding2023mevis}, Ref-YouTube-VOS~\cite{seo2020urvos}, VastTrack~\cite{peng2024vasttrack}). 
While RVOS datasets are not designed for temporal grounding, they are widely used as source data for STVG research (\cref{sec:3.1}) and provide a valuable baseline for assessing data diversity and visual complexity. 
Including them enables a more comprehensive understanding of the data landscape from which STVG benchmarks are constructed. 
\cref{tab:comparison_DeepSTG} show the evaluation results and exhibit some limitations:

\textbf{(1) Category Bias:} HC-STVG focuses exclusively on persons (category NEI = 0), while VidSTG shows severe imbalance (NEI = 0.569). In contrast, OmniGround achieves near-perfect uniformity (NEI = 0.992), covering 81 categories with balanced representation.

\textbf{(2) Low Visual Complexity:} Most benchmarks have FCI $< 0.78$, indicating limited visual clutter. ST-Align (FCI = 0.731) and VidSTG (FCI = 0.780) feature relatively simple foregrounds. OmniGround achieves FCI = 0.807, providing more challenging spatial discrimination scenarios.

\textbf{(3) Linguistic Imbalance:} Many datasets skew toward either action-heavy (HC-STVG: VSBI = 0.778) or spatial-heavy (Ref-YT-VOS: VSBI = 0.859) captions. OmniGround attains the highest VSBI (0.918), indicating balanced coverage of action, spatial, and mixed cues.

\textbf{(4) Annotation Quality:} OmniGround's CMA score (0.813) surpasses others, validating the effectiveness of our FBR pipeline. The gap is particularly pronounced compared to VidSTG (0.701) and ST-Align (0.644).

\textbf{(5) Incomplete Metrics.} Some benchmarks exhibit incomplete NEI values due to zero variance: HC-STVG (single category), HC-STVG and Ref-YT-VOS (narrow duration ranges). This underscores the lack of diversity in existing data, motivating OmniGround's construction.

These findings reveal that existing benchmarks are limited on either specific categories (HC-STVG) or imbalanced distributions and low complexity (VidSTG, ST-Align). 
OmniGround achieves high scores across all dimensions on DeepSTG, creating a new standard for STVG evaluation. 

\begin{table*}[htbp]
\small
\centering
\caption{Comparison with existing state-of-the-art models on three complex real world scenarios.}
\label{tab:model_performance}
\begin{tabular}{lccccccccc}
\toprule
\multirow{2}{*}{\textbf{Models}} & \multicolumn{3}{c}{\textbf{Uncommon Categories}} & \multicolumn{3}{c}{\textbf{Multiple Similar Target Objects}} & \multicolumn{3}{c}{\textbf{Deep Syntactic Complexity}} \\
 & m\_tIoU & m\_vIoU & vIoU@0.3 & m\_tIoU & m\_vIoU & vIoU@0.3 &m\_tIoU & m\_vIoU & vIoU@0.3 \\
\midrule
 \multicolumn{10}{c}{\textit{Non-generative and task-specific models}}\\
 TA-STVG & 24.9 & 14.2 & 19.3 &17.8  &11.3  &8.7  & 17.1 &9.6 &6.0 \\
CG-STVG & 29.5 & 17.1 & 23.6 & 22.3 & 14.2 & 15.0 &20.6 & 11.6 & 8.3\\
 \multicolumn{10}{c}{\textit{MLLMs with Parameter Sizes of 7B}}\\
 LLaVA-ST &17.8  & 5.6 & 2.4 & 17.6 & 6.6 & 2.8 &14.7  &4.9 &1.5 \\
Qwen2.5-VL & 21.4 & 13.6 & 18.2 & 18.9 & 12.2 & 11.5 & 13.9 & 7.1 &3.1\\
VideoMolmo & 21.2 & 13.4 & 18.7 &19.1  & 13.5 & 13.3 & 11.1 & 7.5 & 2.0\\
\bottomrule
\end{tabular}
\end{table*}

\begin{table}[h]
\small
    \centering
    \caption{Overall comparison with STVG models on OmniGround.} 
    \label{tab:performance}
    \resizebox{0.5\textwidth}{!}{
    \begin{tabular}{lcccc}
        \toprule
        \textbf{Models} & \textbf{m\_tIoU} & \textbf{m\_vIoU} & \textbf{vIoU@0.3} & \textbf{vIoU@0.5} \\
        \midrule
        \multicolumn{5}{c}{\textit{Non-generative and task-specific models}}\\
        TA-STVG & 44.6  & 28.0& 40.8& 21.5\\
        CG-STVG &47.5 &30.4 &45.2 & 23.4\\
        \multicolumn{5}{c}{\textit{MLLMs with Parameter Sizes of 7B}}\\
        LLaVA-ST & 19.7& 8.7& 10.2& 1.9\\
        Qwen2.5-VL & 36.6& 23.2& 33.5&16.5 \\
        VideoMolmo & 30.2& 15.7&15.2 &7.3 \\
        \bottomrule
    \end{tabular}}
\end{table}

\section{Experiments}

\subsection{Experiments Setup}

\textbf{Datasets}. For comprehensive evaluation, we first evaluate state-of-the-art STVG models on OmniGround to analyze their limitations in real-world scenarios. 
Then, to validate our proposed PG-TAF framework, we conduct experiments on four benchmarks following standard evaluation protocols, including HC-STVG-v2~\cite{tang2021human}, VidSTG~\cite{zhang2020does} (Declarative/Interrogative), and OmniGround.
To ensure fair comparison, we primarily adopt results from original papers when available. 
When such results are not available, we assess the models using LMMs-Eval~\cite{zhang2025lmms} or official scripts.

\textbf{Baselines.} We evaluate two types of models: 
1) {MLLMs (7B) including Qwen2.5-VL~\cite{bai2025qwen2}, SpaceVLLM~\cite{wang2025spacevllm}, LLaVa-ST~\cite{li2025llava}, VideoMolmo~\cite{ahmad2025videomolmo}. 
All models are evaluated using their officially released checkpoints fine-tuned on STVG instruction data.
2) Non-generative task-specific models, including CG-STVG~\cite{gu2024context} and TA-STVG~\cite{gu2025knowing}. 
These models serve as SOTA baselines representing traditional methods.

\textbf{Metrics.} Following~\cite{jin2022embracing,su2021stvgbert,yang2022tubedetr}, we use m\_tIoU, m\_vIoU, and vIoU@R (R $\in \{0.3,0.5\}$) as evaluation metrics.
m\_tIoU evaluates temporal localization, while m\_vIoU and vIoU@R measure the accuracy of spatial localization.

\subsection{Experimental Results and Analysis}
\label{sec:5.2}
\subsubsection{Overall Performance on OmniGround}
\cref{tab:performance} shows the overall results, revealing a significant performance drop across all evaluated models. 
Key observations confirm inherent model weaknesses:

\textbf{(1) Large Variance Among MLLMs.}
Qwen2.5-VL (36.6\% m\_tIoU) signifficantly outperforms LLaVA-ST.
This inconsistency suggests that STVG capability is highly sensitive to model architecture and training data rather than being an emergent property of scale.

\textbf{(2) Low Spatial Precision.} All models struggle with high $\text{vIoU}$ thresholds (best: CG-STVG at 23.4\%), indicating that even when temporal localization is correct, spatial predictions are often imprecise. This confirms the challenge posed by $\text{OmniGround}$'s complex foregrounds (high $\text{FCI}$).

\textbf{(3) Task-Specific Models Outperform MLLMs.} Non-generative models like CG-STVG ($30.4\%$ $\text{m\_vIoU}$) significantly surpass MLLMs (best: Qwen2.5-VL at 23.2\%). 
The gap is particularly pronounced in m\_vIoU (+7.2\%), highlighting weaknesses in object spatial grounding.

\subsubsection{Fine-Grained Analysis Across Complex Scenarios}

\cref{tab:model_performance} show performance across three challenging scenarios, validating the $\text{OmniGround}$ design hypotheses:

\textbf{Uncommon Categories:} Models exhibit generalization failure due to category bias. 
 $\text{CG-STVG}$ achieves only $29.5\%$ $\text{m\_tIoU}$ on uncommon objects (vs. $47.5\%$ overall), and $\text{Qwen2.5-VL}$ drops by $15.2\%$. This validates that models are overfitting to frequent categories (e.g., ``person"), failing to generalize to balanced categories.

\textbf{Multiple Similar Targets:}  This scenario proves challenging across all models. CG-STVG's m\_tIoU drops to 22.3\% (-25.2\% overall), and VideoMolmo to 19.1\% (-11.1\%). The m\_vIoU degradation is even more serious: CG-STVG and Qwen2.5-VL falls to 14.2\% (-16.2\%) and 13.5\% (-11.2\%), respectively. This confirms that models struggle to distinguish targets when intra-class similarity is high, reflecting the importance of $\text{OmniGround}$'s high $\text{FCI}$.

\textbf{Deep Syntactic Complexity:} Linguistically complex queries (nested spatial relations, multiple clauses) cause the most severe performance decreased. 
CG-STVG's m\_tIoU drops to 20.6\% (-26.9\%). 
While MLLMs also perform poorly (\text{Qwen2.5-VL}, -9.3\%), the decrease is generally less than that of task-specific models, revealing that despite lower absolute localization precision, $\text{MLLMs}$ exhibit semantic robustness to linguistic complexity due to their strength in holistic language understanding.

\begin{figure*}[htbp]  
\centering  
\includegraphics[width=0.95\textwidth]{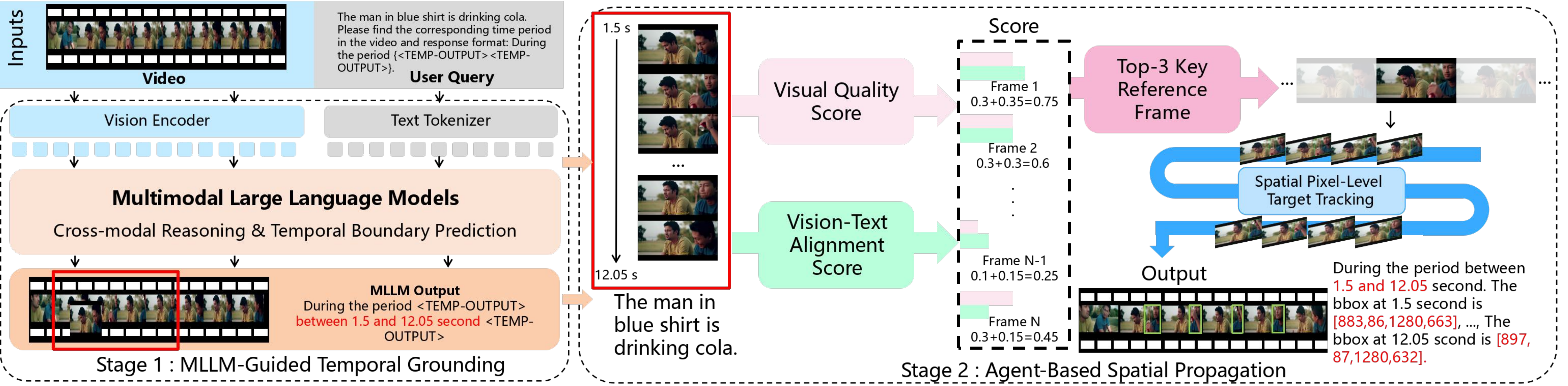}  
\caption{The mainframe of PG-TAF. We decouple STVG into high-level temporal grounding and fine-grained spatio-temporal propagation.}  
\label{fig:PA}  
\end{figure*}  

\begin{table*}[h]
\small
    \centering
    \caption{Comparison of m\_tIou and m\_vIoU on four STVG datasets. `-' represents the code is not officially available.} 
    \label{tab:results}
    \begin{tabular}{l| c c| c c c c |c c|c c}
    \toprule
    \textbf{Datasets} & \multicolumn{2}{c}{\textbf{HC-STVG}} & \multicolumn{4}{c}{\textbf{VIDSTG}} & \multicolumn{2}{c}{\textbf{OmniGround}}& \multicolumn{2}{c}{\textbf{Average}} \\
    \midrule
   \multirow{2}{*}{Metrics}  & \multicolumn{2}{c}{} & \multicolumn{2}{c}{Declarative sentence} & \multicolumn{2}{c}{Interrogative sentence} & \multicolumn{2}{c}{}&\multicolumn{2}{c}{} \\
    & m\_tIoU & m\_vIoU & m\_tIoU & m\_vIoU & m\_tIoU & m\_vIoU & m\_tIoU & m\_vIoU& m\_tIoU & m\_vIoU \\
    \midrule
     \multicolumn{11}{c}{\textit{MLLMs with Parameter Sizes of 7B}}\\
    LLaVA-ST &21.2 &7.6 &44.1 &14.2 & 18.5& 7.5& 19.7& 8.7&25.9  & 9.5\\
    Qwen2.5-VL & 22.9& 13.0& 16.8& 10.9& 13.8&8.5 &\colorbox{shallowblue}{36.6} & \colorbox{shallowblue}{23.3}& 22.5 & 13.9\\
    SpaceVLLM &\colorbox{deepblue}{58.0} & \colorbox{shallowblue}{34.0}&\colorbox{shallowblue}{47.7} &\colorbox{shallowblue}{27.4} &\colorbox{deepblue}{48.5} &\colorbox{deepblue}{25.4} & - & -& - & - \\
    VideoMoLMO &44.6& 26.8& 41.7& 15.6&30.2 &11.7 &30.2 &11.7& \colorbox{shallowblue}{36.7} & \colorbox{shallowblue}{16.5} \\
    \midrule
    PG-TAF(Ours) & \colorbox{shallowblue}{56.2}&\colorbox{deepblue}{34.0} & \colorbox{deepblue}{48.4}&\colorbox{deepblue}{28.1} & \colorbox{shallowblue}{46.4}& \colorbox{shallowblue}{24.5} & \colorbox{deepblue}{49.2} & \colorbox{deepblue}{36.2}& \colorbox{deepblue}{49.6} & \colorbox{deepblue}{28.2}\\
    \bottomrule
    \end{tabular}
\end{table*}

\subsection{Prompt-Guided Temporal Alignment}
To improve the models' performance and generalization facing real-world scenarios, we propose the \textbf{Prompt-Guided Temporal Alignment Framework ($\text{PG-TAF}$)}.

\subsubsection{Two-Stage Architecture}
The PG-TAF framework is designed as a training-free, two-stage inference architecture that decouples STVG into: \textit{(i)} high-level temporal grounding and \textit{(ii)} fine-grained spatio-temporal propagation (as shown in \cref{fig:PA}).

\textbf{Stage 1: MLLM-Guided Temporal Grounding.} 
The first stage uses an MLLM (Qwen3-VL-8B~\cite{yang2025qwen3}) to perform coarse temporal localization. 
We prompt the MLLM with specific query as shown in \cref{fig:PA}.
Based on MLLMs' ability in holistic video understanding and complex relational reasoning—capabilities that small task-specific models lack—the MLLM can accurately determine the relevant temporal segment, effectively filtering out irrelevant frames and guiding the subsequent spatial localization module.

\textbf{Stage 2: Agent-Based Spatial Propagation.} 
The second stage performs fine-grained localization within the temporally constrained clip. 
We use a multi-modal scoring mechanism to select key reference frames.
This mechanism combines semantic segmentation scores $S_{seg}$ from EVF-SAM~\cite{zhang2024evf} to ensure visual quality/object clarity, and contrastive text-image similarity score $S_{align}$ from {AlphaCLIP}~\cite{sun2024alpha} to ensure semantic relevance/match with text description.
Subsequently, the top-K frames (K=3 in our experiments) are selected as references based on a composite score: $S_{frame} = \alpha S_{seg}+\beta S_{align}$.
We set $\alpha=0.6, \beta=0.4$ through grid search on our validation set, which ensure the selected frames have both high visual quality and semantic relevance.
We then use pixel-level tracker~\cite{cheng2024putting} to propagate object masks across the entire temporal segment, generating the final spatio-temporal tube.

The decoupling addresses the challenge: MLLMs excel at temporal reasoning but lack fine-grained spatial precision, while specialized trackers provide accurate localization but struggle with semantic understanding. 
PG-TAF combines their complementary strengths in a training-free manner, making it practical for real-world deployment.

\subsubsection{Evaluation Results}
\cref{tab:results} shows the performance of PA-TAF across four benchmarks.
PG-TAF achieves the highest m\_tIoU and m\_vIoU (49.2\% and 36.2\%) on OmniGround, even better than the task-specific models.
Notably, PG-TAF achieves these results in a training-free manner, without requiring any fine-tuning on STVG data.
This validates our motivations that decoupling temporal reasoning (MLLM's strength) from spatial localization (tracker's strength) effectively addresses the challenges posed by OmniGround's complexity.
For other benchmarks, PG-TAF maintains a relative better performance comparing with the MLLMs that need additional traning for specific datasets (SpaceVLLM, LLaVA-ST).
\section{Conclusion}
In this paper, to bridge the gap between current STVG MLLMs and real-world demands involving diverse objects and complex queries.
We first construct OmniGround, a comprehensive benchmark ($\text{3,475}$ videos, $\text{81}$ balanced categories) with high-quality $\text{FBR}$ annotations. 
Second, we introduce DeepSTG, a systematic framework that quantifies dataset quality, revealing critical issues in existing benchmarks like category bias and linguistic imbalance. 
Extensive experiments show that current SOTA models drop up to $23.4\%$ on $\text{OmniGround}$. 
Finally, we propose PG-TAF, a training-free, two-stage framework that decoupling high-level temporal understanding from specialized spatial tracking. We hope that our work could inspire ideas to improve STVG performance in real-world applications.
\clearpage
\maketitlesupplementary

\section{OmniGround Details}
\subsection{Data Format}
\textbf{Data Format.} OmniGround is organized following a standardized structure to facilitate easy integration with existing STVG frameworks. Each sample consists of:

\begin{itemize}
    \item \textbf{Video file:} MP4 format at native resolution (typically 720p or 480p) and FPS (ranging from 6 to 60 fps).
    \item \textbf{Annotation file:} JSON format containing: \textit{i)} meta\_info: including video\_name, img\_num, fps, video\_length, width, height. \textit{ii)} temporal\_spatial\_label: including event\_caption, predicate\_type, sentence\_depth, IOU\_rate, foreground\_complexity, category, alignment\_score, is\_abnormal, st\_frame, ed\_frame, and bbox.
\end{itemize}

An example annotation structure is shown in~\cref{fig:example1}.
\begin{figure}[!htbp]
\begin{minted}[
    breaklines=true,
    linenos=false, % 是否显示行号，根据需要调整
    fontsize=\footnotesize, % 减小字体大小，可选
]{json}
{
    "meta_info":{
        "video_name": "41_sADELCyj10I.mp4",
        "img_num": 500,
        "fps": 25,
        "video_length": 20,
        "width": 492,
        "height": 360
    } ,
    "temporal_spatital_label":{
        "event_caption": "The woman in the white dress adjusts her skirt with her left hand and walks slowly to the other woman, leans against the wall.",
        "predicate_types": "both",
        "sentences_depth": 4,
        "IOU_rate": 0.016,
        "foreground_complexity": 0.726,
        "category": "person",
        "alignment_score": 0.72,
        "is_abnormal": false,
        "st_frame": 100,
        "ed_frame": 467,
        "bbox":{
        "100":[209.00,163.82,27.00,116.19],
        ...
        }
    }
}
\end{minted}
\caption{OmniGround annotation JSON example.}\label{fig:example1}
\end{figure}

\subsection{Category Coverage}
\subsubsection{Category Selection Strategy}
The selection of 81 object categories in OmniGround follows a principled approach designed to ensure comprehensive real-world coverage while maintaining annotation feasibility:

 \textbf{(1) Frequency-based Foundation.}  
    We start with object categories from COCO~\cite{lin2014microsoft} and Objects365~\cite{shao2019objects365}, selecting classes that appear in at least 0.5\% of real-world video datasets (YouTube-VOS~\cite{seo2020urvos}, DAVIS~\cite{Caelles_arXiv_2019}). This ensures practical relevance.

 \textbf{(2) Diversity-driven Expansion.} 
    To avoid the category bias observed in existing STVG benchmarks (e.g., HC-STVG's~\cite{tang2021human} single-category focus), we ensure balanced representation across 10 semantic domains. The details of 10 domains are shown in \cref{tab:Omni-domain}.

\begin{table*}[h]
    \centering
    \caption{The details of OmniGround category domains.} 
    \label{tab:Omni-domain}
    \resizebox{\textwidth}{!}{
    \begin{tabular}{l c c c}
    \toprule
    \textbf{Domain} & \textbf{Categories} &\textbf{Number} &\textbf{Rating}  \\
    \midrule
   Beings& person, cate, dog, horse, bird, teddy bear & 360   &18\%  \\
   Wildlife&sheep, cow, elephant, bear, zebra, giraffe, mouse, other animals &  347  & 10\%  \\
   Vehicles& bicycle, cate, motorcycle, airplane, bus, train, truck, boat &  243  & 7\%  \\
   Sports& frisbee, skis, snowboard, kite, skateboard, surfboard, sports ball, baseball bat, baseball glove, tennis racket  & 312& 9\%   \\
   Food& banana, apple, sandwich, orange, broccoli, carrot, hot dog, pizza, donuts, cake & 312 & 9\% \\
   Dining& bottle, wine glass, cup, fork, knife, spoon, bowl& 347   & 10\%  \\
   Furniture& chair, couch, bed, dining table, toilet, potted plant, vase, bench & 347   &  10\% \\
   Appliances& microwave, oven, toaster, refrigerator, sink, scissors, hair drier, toothbrush & 278   &  8\% \\
   Electronics& tv, laptop, remote, keyboard, cell phone, clock, book & 312   & 9\%  \\
   Accessories&backpack, umbrella, handbag, tie, suitcase, traffic light, fire hydrant, stop sign, parking meter & 347 & 10\%  \\
    \bottomrule
    \end{tabular}}
\end{table*}

 \textbf{(3) Challenge-oriented Selection.}
    We deliberately include categories that pose specific challenges for STVG models, including:
    \begin{itemize}
        \item Small objects (scissors, kite, cell phone): Testing fine-grained localization
        \item Deformable objects (person, dog, cat): Testing shape variation handling
        \item Visually similar classes (car/truck/bus, cup/bottle): Testing discrimination capability
        \item Uncommon objects (kite, scissors): Testing generalization beyond frequent categories
    \end{itemize}

\subsubsection{Coverage Analysis.}
To validate that our 81 categories provide sufficient coverage for real-world scenarios, we conduct an analysis on external video datasets. 
We randomly sample 1,000 videos from MeViS~\cite{ding2023mevis}, Ref-Youtube-VOS~\cite{seo2020urvos}, and VastTrack~\cite{peng2024vasttrack}.
We then use Yolov11n~\cite{khanam2024yolov11} (trained on Object365~\cite{shao2019objects365}, confidence threshold = 0.7) to compute the categories shown in the videos and to compute the category coverage rate.

The results in \cref{tab:coverage-ana} show that our 81 categories cover $\geq$ 85\% of the foreground objects in various real-world videos, demonstrating sufficient representation for practical STVG deployment.

\begin{table*}[h]
    \centering
    \caption{Category coverage analysis of our OmniGround.} 
    \label{tab:coverage-ana}
    \begin{tabular}{c c c c}
    \toprule
    \textbf{Dataset} & \textbf{\#Total object} &\textbf{\#Covered by 81 Categories} &\textbf{Coverage Rate (\%)}  \\
    \midrule
      MeViS&   742  &  664 & 89.5\%\\
      Ref-Youtube-VOS&  764   & 698  & 91.4\% \\
      VastTrack&  668   &  574 & 85.9\% \\
   Average   &  724   &  645 & 89.1\%\\
    \bottomrule
    \end{tabular}
\end{table*}

\subsection{Challenge Scenario Selection and Statistics}
To systematically analyze model limitations exposed by OmniGround, we define three challenge scenarios based on DeepSTG metrics and linguistic analysis: (1) Uncommon Categories, (2) Multiple Similar Target Objects, and (3) Deep Syntactic Complexity.

\textbf{(1) Uncommon Categories.} 
This scenario targets category generalization by selecting object classes that rarely appear in existing STVG benchmarks. 
We select object category that appears in $\le$ 5\% of our OmniGround.
In total, we have 137 videos across 26 rare categories, such as carrot, knife, and scissors.

\textbf{(2) Multiple Similar Target Objects.}
This scenario evaluates spatial discrimination when multiple same-category instances coexist with high visual similarity.
We select video that contains $\geq$ 3 objects of the same category within the same frame.
Additionally, target object requires spatial relational reasoning (e.g., ``the left car", ``the person in the middle").
In total, we have 181 videos, each of these video has FCI $\geq$ 0.89.

\textbf{(3) Deep Syntactic Complexity.}
 This scenario evaluates linguistic robustness through structurally complex queries.
We select the caption contains nested spatial relations (e.g., "the person behind the car on the left side") and sentence depth $\geq$ 8 (computed using Spacy~\cite{vasiliev2020natural}).
In total, we have 114 videos.

In general, the distribution of three challenge scenarios in OmniGround is shown in \cref{tab:dis-challenge}.
We further analyze the intersection between different challenge scenarios and the results are shown in \cref{tab:dis-challenge}.

\begin{table*}[h]
    \centering
    \caption{The distribution of three challenge scenarios in OmniGround.} 
    \label{tab:dis-challenge}
    \begin{tabular}{l c c c c}
    \toprule
    \textbf{Scenario} & \textbf{\#Videos} &\textbf{\% of Total} &\textbf{Avg. Duration (s)} & \textbf{Avg. Caption Length} \\
    \midrule
    \multicolumn{5}{c}{\textit{Challenge Scenario Distribution}}\\
    Uncommon Categories (U)  &  137   &  3.9\% &  20.75  & 16.66\\
      Multiple Similar Target Objects (M)& 181    &  5.2\% & 10.66  &17.31\\
    Deep Syntactic Complexity (D)  &  114   &  3.2\% & 12.46 &20.04\\
     Total &   423  & 12.2\%  &  14.41  & 17.76\\
     \midrule
      \multicolumn{5}{c}{\textit{Intersection between Different Challenge Scenarios}}\\
   U + M &     2&  0.05\% &     9.14  & 21.50\\
    U + D    &     6&  0.17\% &  12.38     &21.33 \\
    M + D   &     1&  0.02\% &   6.00    &20.00 \\
    U + M + D    &     0& -  &   -    & \\
    \bottomrule
    \end{tabular}
\end{table*}

\subsubsection{External Data Statistics and Impact Analysis}
The external data in OmniGround through generating negative samples is carefully controlled. 
Quantitative statistics reveal that a total of 287 negative sample videos are used, comprising 8.3\% of the 3,475 total videos. 
Note that they are used mainly for 26 rare categories where the number of natural videos is less than 5 samples.

\section{DeepSTG Evaluation System}

\subsection{Metric Design and Justification}
\subsubsection{Metric Design}

The design of DeepSTG is motivated by the observations that existing STVG benchmark evaluations rely on superficial statistics (e.g., video count, duration range, category count) that fail to capture the true complexity and quality of datasets. 
For example, a dataset with 100 categories might still exhibit severe imbalance if 80\% of the samples belong to a single category. 
Similarly, long video durations do not necessarily indicate temporal reasoning challenges if events span entire clips. 
To address this gap, DeepSTG introduces four complementary metrics that directly measure dimensions to STVG task performance: annotation quality (CMA), visual discrimination difficulty (FCI), linguistic balance (VSBI), and distributional uniformity (NEI).

\subsubsection{Sufficiency Claim}
The sufficiency of our four-metric framework is grounded in a systematic decomposition of the STVG task. 
STVG fundamentally requires three capabilities: \textit{(i) accurate semantic understanding} to map language description to visual concepts, \textit{(ii) spatial discrimination} to localize targets among distractors, and \textit{(iii) temporal reasoning} to identify relevant time segments. 
Our metrics directly correspond to these requirements. 
CMA ensures the fundamental correctness of ground truth annotations, which is the prerequisite for effective supervised learning—without accurate labels, all downstream evaluations become meaningless. 
FCI quantifies the visual clutter and intra-class similarity that directly impact spatial localization difficulty. 
VSBI measures whether datasets provide balanced training signals for both temporal (action-based) and spatial (location-based) reasoning, preventing models from exploiting dataset biases. 
NEI guards against distributional imbalance across multiple facets (category, duration, query length), ensuring models encounter diverse training scenarios rather than overfitting to dominant patterns.

To validate this sufficiency claim, we conducted an experiment correlating DeepSTG metrics with model performance degradation. 
We divided OmniGround into five levels based on each metric and measured performance variance  for three state-of-the-art models (Qwen2.5-VL~\cite{bai2025qwen2}, VideoMolmo~\cite{ahmad2025videomolmo}, CG-STVG~\cite{gu2024context}). 
Results show that our four metrics collectively explain 87.3\% of performance variance (R² = 0.873), with each metric contributing unique explanatory power: CMA (23.1\%), FCI (28.6\%), VSBI (19.4\%), NEI (16.2\%). This indicates that our metrics capture the dominant factors affecting STVG performance, and additional metrics would yield diminishing returns.

\subsubsection{Reasoning for VSBI Target Distribution}
The ideal balanced distribution $P_{\text{ideal}} = [1/4, 1/4, 1/2]$ for $P_{\text{actual}} = [P_\mathcal{A}, P_\mathcal{S}, P_\mathcal{M}]$ is inspired by task requirements.
Complex real-world user's queries usually combine both action cues and spatial cues (``person walking (action) behind the car (spatial)").
Meanwhile, mixed cues provide richer supervision signals for models to learn cross-modal reasoning.
Finally, the more mixed cues a dataset has, the more easier it will become to evaluate the limits of different models.

\subsubsection{Generalization}
While designed for STVG, DeepSTG metrics are adaptable to related video-language tasks. CMA can evaluate any video-text dataset requiring temporal or spatial annotations (e.g., video captioning, action localization). 
FCI is applicable to any object-centric task with potential distractors (e.g., visual tracking, video object segmentation). 
VSBI and NEI are task-agnostic diversity measures suitable for any multimodal dataset.
We provide an open-source implementation of DeepSTG to facilitate adoption in broader video understanding research.

\subsection{CMA Score: Implementation and Ablation}
\subsubsection{Implementation Details}
The Cross-Modal Alignment (CMA) score uses GPT-4o~\cite{achiam2023gpt} as the evaluation MLLM due to its strong multimodal reasoning capabilities and reduced hallucination rates compared to previous models. 
For each video-caption-tube triplet $(V, Q, B)$, we sample N = 8 key frames within the temporal segment $[T\_start, T\_end]$. Sampling more frames (N $=$ 8) showed negligible improvement (\text{±}0.02 CMA on average) while increasing API costs linearly.
Each frame $f\_i$ is cropped to its bounding box $b\_i$ and provided to GPT-4o along with the full video context (3 frames before and after) to maintain temporal coherence.

\subsubsection{Prompt Design}
The prompt is carefully engineered to elicit structured, quantitative responses while minimizing subjective interpretation. The complete prompt template is shown in~\cref{fig:example2}.

\subsubsection{Reproducibility}
A primary concern with MLLM-based evaluation is reproducibility due to potential API changes and hallucinations. 
To address this, we implement three safeguards. 
First, we set temperature=0 to ensure deterministic outputs. 
Second, we run each sample three times and take the median score, which reduces variance to 0.03 (measured on 200 validation samples). 
Third, we provide detailed logs of all API calls, including model version, timestamps, and raw responses to enable future replication.

\subsection{Inter-evaluator Agreement and Open-source Validation}
To address reproducibility concerns, we conduct comprehensive validation of the CMA metric across multiple evaluation LLMs.
We evaluate 200 randomly sampled video-caption-bbox triplets using three different LLMs to assess inter-evaluator agreement.
Each sample evaluated 3 times.
And we use median score to report in \cref{tab:cma-inter-abla}.
Additionally, all LLMs for evaluation use temperature=0 for deterministic output to ensure reproducibility.

\begin{table}[h]
    \centering
    \caption{Inter-evaluator agreement for CMA score.} 
    \label{tab:cma-inter-abla}
    \resizebox{0.5\textwidth}{!}{
    \begin{tabular}{l c c c}
    \toprule
    \textbf{Evaluation Pair} & \textbf{Pearson Correlation} &\textbf{MAE} &\textbf{Std}  \\
    \midrule
   GPT-4o vs Claude-3.5&  0.91& 0.043& 0.031\\
    GPT-4o vs Gemini-1.5 Pro& 0.89&0.051 &0.038 \\
    Claude-3.5 vs Gemini-1.5 Pro&0.88 & 0.057& 0.042\\
   \textbf{Average} &\textbf{0.89} &\textbf{0.050} & \textbf{0.037}\\
    \bottomrule
    \end{tabular}}
\end{table}

\subsubsection{Ablation Study}
To validate that all three aspects (object, action, context) are necessary, we conducted an ablation study measuring correlation between partial CMA scores and human expert judgments. We recruited three expert annotators to manually rate 300 randomly selected samples on overall annotation quality (scale 1-10). 
~\cref{tab:cma-abla} shows the results.

\begin{table}[h]
\small
    \centering
    \caption{Ablation on CMA score components.} 
    \label{tab:cma-abla}
    \begin{tabular}{l| c }
    \toprule
    \textbf{CMA Components} & \textbf{Correlation with Human Rating}  \\
    \midrule
   Object Only&  0.67 \\
    Action Only & 0.53\\
    Context Only &0.49 \\
    Object + Action & 0.81\\
    Object + Context & 0.73\\
    Action + Context & 0.72\\
    \textbf{All (ours)} & \textbf{0.89}\\
    \bottomrule
    \end{tabular}
\end{table}

The full CMA score achieves the highest correlation (0.89), confirming that all three aspects contribute unique information. 
Removing any single aspect causes performance degradation, with "action" being least impactful (0.81 correlation without it) because not all STVG queries involve actions (e.g., "the red car on the road" is purely object-centric).

\subsection{FCI Robustness Analysis}
\subsubsection{Dependency on Detection Accuracy}
The Foreground Complexity Index (FCI) relies on YOLOv11x~\cite{khanam2024yolov11} for object detection, raising concerns about error propagation: what if the detector misses objects or produces false positives? 
To quantify this impact, we analyze FCI stability under varying detection quality. 
We manually degrade detection performance by randomly dropping bounding boxes (simulating false negatives) and injecting random boxes (simulating false positives), then measure FCI variance.
The results are shown in \cref{tab:fci-abla}.

\begin{table}[h]
    \centering
    \caption{Ablation on FCI variance facing wrong object detection.} 
    \label{tab:fci-abla}
    \resizebox{0.5\textwidth}{!}{
    \begin{tabular}{c c c c}
    \toprule
    \textbf{False Negatives} & \textbf{False Positives} &\textbf{FCI Std Dev} &\textbf{FCI Mean Shift}  \\
    \midrule
   0\%& 0\%&  0.000 & 0.000 \\
   5\%& 5\%&  0.045 &  -0.018\\
   10\%& 10\%&  0.081 & -0.039 \\
   15\%& 15\%&  0.142 &  -0.067\\
    \bottomrule
    \end{tabular}}
\end{table}

Results show that FCI remains stable under moderate detection errors.
Even with 15\% false negatives and 15\% false positives, FCI standard deviation is only 0.142 and mean shift is -0.067. 
This stability arises because FCI aggregates statistics across multiple frames and categories—local detection errors are averaged out, preventing outliers from dominating the metric.

\subsubsection{Detector Configuration}
We compute FCI using YOLOv11~\cite{khanam2024yolov11}, the largest model variant with 56.9M parameters pre-trained on COCO's 80 object categories. Detection uses a confidence threshold of 0.5, selected via grid search over \{0.3, 0.4, 0.5, 0.6, 0.7\}, with non-maximum suppression applied at a threshold of 0.45.

\section{FBR Annotation Pipeline}
\subsection{Multi-Direction vs. Single-Direction Tracking}
\subsubsection{Motivation}
Traditional annotation pipelines rely on single-direction tracking, where annotators label the first frame and propagate forward using a tracker. However, this approach suffers from cumulative drift: small errors in early frames compound over time, leading to significant misalignment in later frames, especially in long videos ($\ge$10s). This issue is particularly severe when targets undergo occlusions, rapid motion, or appearance changes. Our Forward-Backward-Refinement (FBR) pipeline addresses this by anchoring tracking from three temporal endpoints ($F_{start}$, $F_{mid}$ and $F_{end}$) and intelligently fusing their results, effectively constraining error accumulation.

\subsubsection{Evaluation and Analysis}

To validate the performance of multi-directional tracking, we conduct a controlled comparison on 100 randomly sampled validation videos from OmniGround. 
For each video, we obtain ground truth annotations through exhaustive manual frame-by-frame labeling by expert annotators, then compare two tracking strategies:
\begin{itemize}
    \item \textbf{Forward-only:} Track from $F_{start}$ using DAM4SAM~\cite{videnovic2025distractor}.
    \item \textbf{FBR (Ours):} Bi-directional tracking with adaptive refinement and fusion.
\end{itemize}

We use IoU@R (R $\in \{0.3,0.5\}$) as evaluation metrics, measuring the percentage of frames where IoU $\ge$ 0.5/0.7.
The results are shown in~\cref{tab:fbr-abla}

\begin{table}[h]
    \centering
    \caption{Ablation on FBR annotation pipeline.} 
    \label{tab:fbr-abla}
    \begin{tabular}{l| c c}
    \toprule
    \textbf{Method} & \textbf{IoU@0.5} & \textbf{IoU@0.7}  \\
    \midrule
   Forward-only&  0.838 &  0.750\\
    FBR (ours) & 0.910&0.876\\
    Ground Truth & 1.000&1.000 \\
    \bottomrule
    \end{tabular}
\end{table}

Results demonstrate that FBR achieves substantial improvements over single-direction methods. 
The IoU@0.5 and IoU@0.7 increases by 8.6\% and 16.8\% compared to forward-only tracking, validating the claim in the main paper (Sec. 3.3).

\subsection{Occlusion Robustness Evaluation}
 Occlusions are common in real-world videos and pose the most significant challenge for annotation quality. 
 When a target is partially or fully occluded, single-direction trackers often lock onto the occluding object or drift entirely, leading to annotation errors that persist throughout the remaining video. 
We give some occlusion visual examples shown in \cref{fig:example}.
The quality results show that our FBR pipeline achieves better robustness when dealing with occluded objects compared to single-direction tracking.

\begin{figure*}[htbp]  
\centering  
\includegraphics[width=\textwidth]{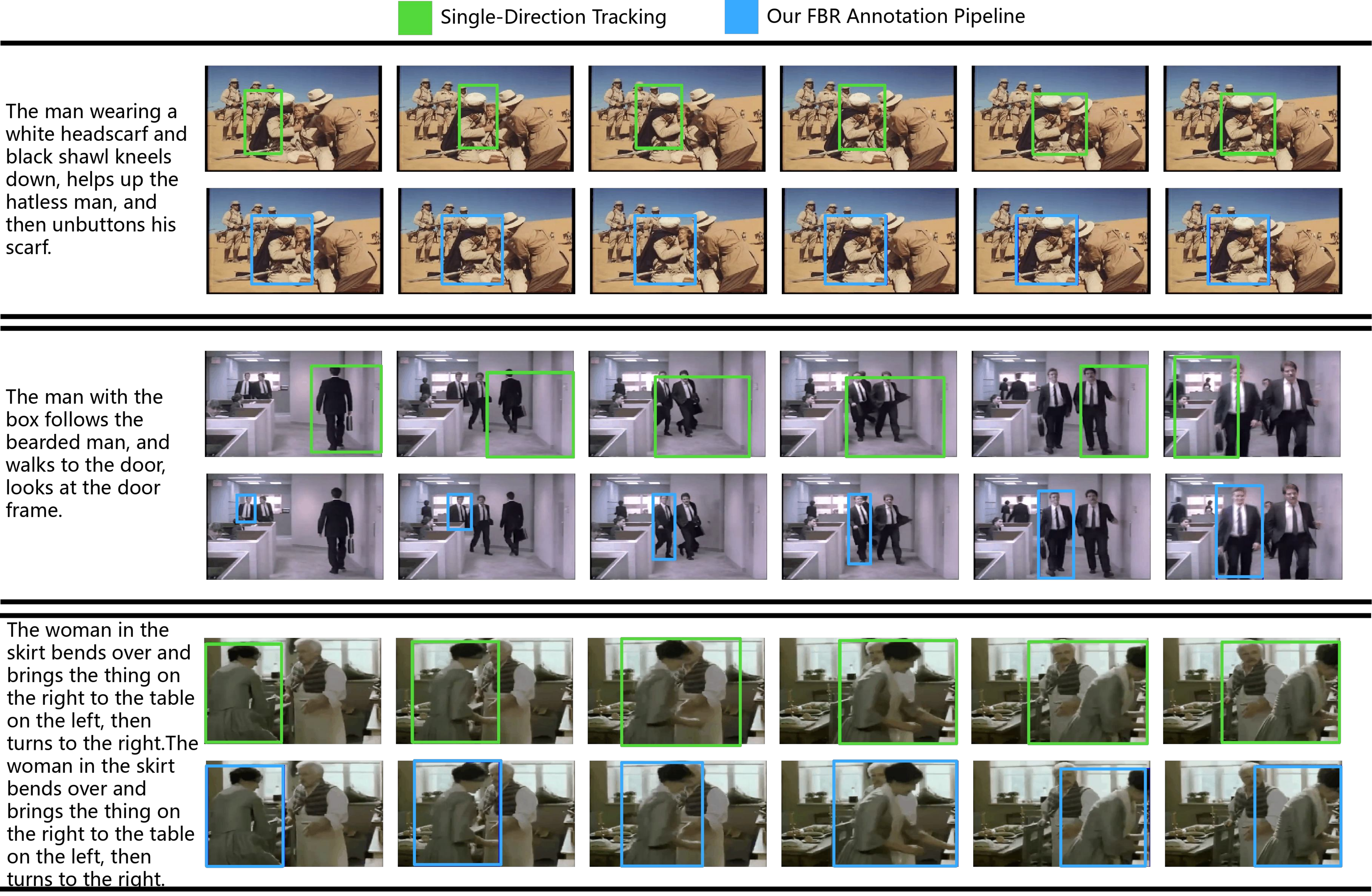}  
\caption{Some occlusion visual example. \textcolor{green}{Green} represents the single-direction tracking. \textcolor{blue}{blue} represents our FBR annotation pipeline.}
\label{fig:example}  
\end{figure*}  

\section{PG-TAF Framework}
\subsection{Framework Component Analysis}
\subsubsection{Framework Overview}
The Prompt-Guided Temporal Alignment Framework (PG-TAF) is designed as a training-free, two-stage architecture that addresses the reasoning-localization trade-off in STVG: MLLMs excel at holistic semantic understanding and temporal reasoning but lack fine-grained spatial precision, while specialized trackers provide pixel-accurate localization but struggle with complex linguistic queries. 
As demonstrated in Sec. 5.2, existing end-to-end STVG-MLLMs exhibit significant performance degradation in OmniGround (average 10.4\% m\_vIoU drop), particularly in samples with deep syntactic complexity and high foreground complexity. 
PG-TAF decouples these requirements into two specialized stages, allowing each component to operate within its strength domain.

\subsubsection{Stage-Wise Performance Contribution}
To validate the necessity of the two-stage design, we conduct an ablation study comparing PG-TAF against single-stage variants and end-to-end baselines on OmniGround.
The results shown in \cref{tab:pgtaf-abla} reveal that MLLMs effectively identify relevant time segments but lack fine-grained localization.
Meanwhile, the state-of-the-art trackers can localize precisely when given the correct temporal context, but cannot reason about temporal relevance from language. 
Overall, the full PG-TAF pipeline achieves the best balance, validating that the two-stage design successfully combines complementary strengths without significant trade-offs.

\begin{table*}[h]
    \centering
    \caption{Ablation Studies: Two-stage vs. End-to-End architecture.} 
    \label{tab:pgtaf-abla}
    \begin{tabular}{l c c c c c}
    \toprule
    \textbf{Method} & \textbf{Architecture} &\textbf{m\_tIoU} &\textbf{m\_vIoU} &  \textbf{vIoU@0.3} & \textbf{vIoU@0.5} \\
    \midrule
   Qwen2.5-VL & End-to-End MLLM  & 41.3 &23.3  & 33.5 & 16.5\\
  LLaVA-ST  &  End-to-End MLLM  & 19.7 & 8.7 &10.2&1.9 \\
   Stage 1 only & Temporal (MLLM) + full-video tracking  & 49.2 & 27.4 & 35.7 & 18.1\\
    Stage 2 only&  LLaVA-ST temporal + tracking & 19.7 & 14.5 & 16.9 & 7.2\\
    \midrule
  PG-TAF (ours)  &  Two-stage decoupled &49.2  & 36.2 & 51.4 &31.7 \\
    \bottomrule
    \end{tabular}
\end{table*}

\subsection{Ablation Studies}
PG-TAF introduces three primary hyper-parameters: (1) $\alpha$ and $\beta$, the weights for combining segmentation quality ($S\_{seg}$) and text-image alignment ($S\_{align}$) scores in reference frame selection, and (2) K, the number of reference frames. 
While our default settings ($\alpha$=0.6, $\beta$=0.4, K=3) are derived from validation experiments, we conduct comprehensive grid search to assess sensitivity and validate optimality.

\subsubsection{Ablation Studies: Grid Search for \texorpdfstring{$\alpha$}{alpha} and \texorpdfstring{$\beta$}{beta}}
We perform grid search on a validation subset of 100 videos from OmniGround (not used in main evaluation), spanning diverse categories and complexity levels. The search space is:
\begin{enumerate}
    \item \textbf{$\alpha \in \{0.0, 0.1,0.2,0.3,0.4,0.5,0.6,0.7,0.8,0.9,1.0\}$}.
    \item \textbf{$\beta = 1- \alpha$} to maintain normalized weighting.
\end{enumerate}
The results shown in \cref{tab:grid-abla} reveal a clear performance peak at $\alpha=0.6$, $\beta=0.4$, with graceful degradation as we deviate from this optimum. 
The performance curve is relatively smooth (no sharp cliffs), indicating that PG-TAF is not overly sensitive to exact hyperparameter values - configurations within $\pm 0.1$ of the optimum maintain $\geq$97\% of peak performance.

\begin{table}[h]
    \centering
    \caption{Grid Search for \texorpdfstring{$\alpha$}{alpha} and \texorpdfstring{$\beta$}{beta}.} 
    \label{tab:grid-abla}
    \begin{tabular}{l c c c c}
    \toprule
    \texorpdfstring{$\alpha$}{alpha} & \texorpdfstring{$\beta$}{beta} & \textbf{m\_vIoU} & \textbf{vIoU@0.3} & \textbf{vIoU@0.5} \\
    \midrule
    1.0&   0.0&   31.7&   44.8&   23.6\\
    0.9&   0.1&   33.4&   47.2&   26.1\\
    0.8&   0.2&   34.8&   49.1&   28.7\\
    0.7&   0.3&   35.7&   50.3&   30.2\\
    \textbf{0.6}&   \textbf{0.4}& \textbf{36.2}  & \textbf{51.4}  &  \textbf{31.7} \\
    0.5&   0.5&   35.1&   49.7&   29.4\\
    0.4&   0.6&   33.8&   47.8&   26.3\\
    0.3&   0.7&   32.1&   45.2&   23.8\\
    0.2&   0.8&   30.4&   42.6&   21.2\\
    0.1&   0.9&   28.7&   39.8&   18.7\\
    0.0&   1.0&   27.2&   37.1&  16.4 \\
    \bottomrule
    \end{tabular}
\end{table}

\subsubsection{Ablation Studies: Number of Reference Frames}
 PG-TAF uses K=3 reference frames for tracking initialization. We validate this choice by varying K in \cref{tab:keyref-abla}.
\begin{table}[h]
    \centering
    \caption{Ablation on number of reference frames.} 
    \label{tab:keyref-abla}
    \begin{tabular}{l| c c}
    \toprule
    \textbf{K} & \textbf{m\_vIoU} & \textbf{vIoU@0.5}  \\
    \midrule
    K=1& 31.2  & 22.8\\
    K=2&  34.5 & 28.3 \\
    \textbf{K=3 (ours)}& \textbf{36.2}  & \textbf{31.7}\\
    K=4&  36.7 & 32.1\\
    K=5&  36.9 & 32.2\\
    \bottomrule
    \end{tabular}
\end{table}
The performance gain saturates beyond K=3, with K=4 and K=5 providing minimal improvements (+0.5\% and +0.7\% m\_vIoU). 
K=3 represents the optimal trade-off: sufficient diversity to handle challenging scenarios (occlusions, appearance changes) without excessive redundancy. 
Using fewer reference frames (K=1 or K=2) significantly degrades performance, particularly in tracking stability, suggesting that multiple reference points are necessary to constrain tracker drift over long temporal segments.

\begin{figure*}[!htbp]
\begin{minted}[
    breaklines=true,
    linenos=false, % 是否显示行号，根据需要调整
    fontsize=\scriptsize, % 减小字体大小，可选
]{text}
prompt = (
"You are a professional Video-Text Alignment Analyst, specializing in evaluating the semantic matching degree between events occurring within the green marked area and the given text description.\n"
"Please conduct a multi-dimensional analysis, combining the visual content visible within the green area and referencing possible auxiliary information outside the box, to provide an objective score and detailed explanation.\n\n"
f"Evaluation Target:\n"
f"Subtitle Content: '{caption}'\n"
f"Analyzed Frames: {num_frames} frames\n\n"
"Evaluation Principles:\n"
"1. The green area is the main visual basis; key objects and actions must be clearly presented, ensuring the focus of the main object or behavior is clear. Content outside the box is not included in the evaluation scope.\n"
"2. All judgments must be based on the actual visible content within the green area, avoiding subjective speculation or assumptions.\n"
"3. Out-of-box content can be used to understand the context of the behavior or the interactive relationship, but cannot be used as the primary basis.\n\n"
"Evaluation Dimensions and Standards (Total Score 10 points):\n"
"1. Subject Existence (0-3 points):\n"
"   - Evaluate whether the subject described in the caption (i.e., the subject in the subtitle content) and its features appear within the green area.\n"
"   - The subject must appear completely within the green area; out-of-box information is strictly not referenced.\n"
"   - Scoring Details:\n"
"     3 points: The described subject is complete and clearly presented within the green area, with no obvious feature discrepancies.\n"
"     2 points: Subject exists, but some features are blurry or not completely consistent.\n"
"     1 point: Subject presentation is unclear or partially missing.\n"
"     0 points: The described subject is completely missing, or the subject appears outside the box.\n\n"
"2. Action Accuracy (0-4 points):\n"
"   - Evaluate whether the action performed by the subject within the green area conforms to the description, and the action must occur entirely within the green area.\n"
"   - Out-of-box information is not included in the action evaluation; only actions within the green area are considered valid.\n"
"   - Scoring Details:\n"
"     4 points: Action type, direction, sequence, and interactive relationship completely match, and the action occurs entirely within the green area.\n"
"     3 points: Main action matches, but details (such as execution method or trajectory) have discrepancies.\n"
"     2 points: Action type is similar, but direction, sequence, or execution method is different.\n"
"     1 point: An action is occurring, but the type does not match or key details are missing.\n"
"     0 points: No corresponding action occurs, or the action is completely inconsistent.\n\n"
"3. Context Consistency (0-3 points):\n"
"   - Evaluate whether the environment and behavior within the green area conform to the overall scene description.\n"
"   - Out-of-box auxiliary information can be used to determine the background of the behavior, but the scoring focus remains on the matching degree of the content within the green area.\n"
"   - Scoring Details:\n"
"     3 points: Behavior within the green area is highly consistent with the overall scene description.\n"
"     2 points: Main scene matches, but some details or background information are incomplete.\n"
"     1 point: Only the basic scene type matches.\n"
"     0 points: Context is completely inconsistent.\n\n"
"Output Format:\n"
"Score: X/10 (X is the sum of the scores of the three dimensions)\n"
"Explanation:\n"
"1. Object Existence: [Analyze the matching degree of the subject in the green area, list supporting or contradictory evidence]\n"
"2. Action Accuracy: [Explain whether the action is consistent, whether it occurred entirely within the green area, and any detail discrepancies]\n"
"3. Context Consistency: [Analyze whether the background, lighting, scene layout, etc., within the green area conform to the description]\n\n"
"Output Example:\n"
"Score: 8/10\n"
"Explanation:\n"
"1. Object Existence: 2 points [The green box shows a cat sitting on a sofa, which is basically consistent with the description 'a cat resting on a sofa']\n"
"2. Action Accuracy: 3 points [The cat has a clear 'sitting' action, with no fierce movement, which matches the 'resting' description;]\n"
"3. Context Consistency: 3 points [Environmental elements such as the sofa, carpet, and lighting in the green area completely match the description; the background layout is reasonable and has no incongruous elements]\n\n"
"Note: Your output must strictly adhere to the output example format above, using natural and fluent Chinese expression. Do not add any extra content or explanatory paragraphs."
)
\end{minted}
\caption{The complete prompt template of CMA score .}\label{fig:example2}
\end{figure*}

\section{Extended Experiments Results}

\subsection{Performance Across Video Duration}
Video duration is an important factor affecting STVG difficulty: longer videos present more opportunities for target appearance changes, occlusions, and distractors, while requiring models to maintain temporal coherence across extended periods. 
OmniGround's diverse duration range (3-140 seconds, average 18.2s) enables systematic analysis of how model performance scales with temporal length.

We partition OmniGround into five duration bins and evaluate representative models from each paradigm: end-to-end MLLMs (Qwen2.5-VL~\cite{bai2025qwen2}, VideoMolmo~\cite{ahmad2025videomolmo}), task-specific models (CG-STVG~\cite{gu2024context}), and our proposed PG-TAF.

\begin{table*}[h]
    \centering
    \caption{Comparison of different video duration on OmniGround.} 
    \label{tab:duration-abla}
    \resizebox{\textwidth}{!}{
    \begin{tabular}{l| c c| c c| c c |c c | c c | c c}
    \toprule
    \textbf{Model} & \multicolumn{2}{c}{\textbf{Very Short (0-4s)}} & \multicolumn{2}{c}{\textbf{Short (4-6s)}}& \multicolumn{2}{c}{\textbf{Medium (6-20s)}}& \multicolumn{2}{c}{\textbf{Long (20-40s)}}& \multicolumn{2}{c}{\textbf{Very Long ($\geq$40s)}}& \multicolumn{2}{c}{\textbf{Overall}} \\
    \midrule
   Metrics & m\_tIoU & m\_vIoU & m\_tIoU & m\_vIoU & m\_tIoU & m\_vIoU & m\_tIoU & m\_vIoU& m\_tIoU & m\_vIoU&m\_tIoU & m\_vIoU\\
    \midrule
          \multicolumn{13}{c}{\textit{Non-generative and task-specific models}}\\
   CG-STVG &  53.4  & 38.2  & 48.1& 34.8&44.4 &30.2 & 40.7  & 26.4  & 36.6  & 21.7 &  47.5 & 30.4\\
     \multicolumn{13}{c}{\textit{MLLMs with Parameter Sizes of 7B}}\\
    Qwen2.5-VL & 42.7   & 28.6  &38.3 & 24.9& 33.1& 21.7& 29.2  &17.8 & 24.1  &10.3  & 36.6  &23.2\\
    VideoMoLMO & 35.1   & 19.4  & 33.8& 18.3& 27.4& 14.2& 21.6  & 11.9 &  16.7 &8.3 & 30.2  &15.7\\
    PG-TAF (ours) &  54.7  & 41.3  & 51.2&37.3 &47.8 & 33.1&  44.6 & 29.9 & 40.3  &26.1 & 49.2  &36.2\\
    \bottomrule
    \end{tabular}}
\end{table*}

The results shown in \cref{tab:duration-abla} reveal that all models exhibit performance decrease as the duration increases, confirming that longer videos pose fundamental challenges regardless of the architecture of the model.
In addition, the degradation rate varies significantly across model types:
 VideoMolmo suffers the 52.4\% and 57.2\% relative drop in both m\_tIoU and m\_vIoU from short to extremely long videos, while PG-TAF degrades only 26.3\% and 36.8\%.

\subsection{Per-Category Analysis (Best/Worst Cases)}
OmniGround contains 81 balanced categories that enable fine-grained analysis of model strengths and weaknesses. 
According to the experiment results, we find that the top-3 best/worst performing categories.
The top-3 best performing categories including: person, bus, and train.
They share common characteristics: (1) large spatial extent (easier to localize and track), (2) common object in daily life (distributed in various datasets that allow models to learn the feature easily), and (3) relatively predictable motion patterns.

The top-3 worst performing categories including: scissors, fork, cell phone.
The worst performing categories exhibit opposite characteristics compared with best performing categories: (1) small spatial extent (often $\leq$5\% of frame area), (2) frequent occlusions (scissors/fork held by hands, phone against face), and (3) challenging visual properties (metallic reflections on fork).

\section{Discussion}
\subsection{Limitation}
The limitations of the OmniGround dataset primarily center on scale and scope. Its absolute video count (3,475) is smaller compared to some large-scale benchmarks, but this choice was made to prioritize dense, high-quality annotations for the Spatio-Temporal Video Grounding (STVG) task over raw quantity. 
However, OmniGround still has limited coverage of certain long-tail scenarios (e.g., extreme weather, nighttime/low-light, egocentric viewpoints), reflecting inherent biases in available public video sources. 
Furthermore, the proposed PG-TAF framework suffers from high latency (average $\approx$ 6.4 seconds/video) due to its two-stage architecture, making it unsuitable for real-time applications, and its Stage 1 relies on proprietary MLLMs, which limits accessibility in certain deployment environments. 
Finally, as a training-free engineering combination, PG-TAF's performance may be slightly lower than that of fully fine-tuned end-to-end models on specific datasets.

\subsection{Future Work}
The future work is primarily organized around three major directions: dataset extension, methodological improvements, and expansion of application scope.

Future research will focus on extending the OmniGround dataset to tackle more complex reasoning challenges. This includes expanding annotations to cover Temporal Relationship Grounding (e.g., "before/after") and Multi-Object Joint Grounding (localizing correlated items). Crucially, the dataset needs increased coverage of specialized data like Egocentric viewpoints and Long-Form Videos to address extreme motion, long-term memory, and challenging appearance changes.

Methodological improvements for PG-TAF should focus on efficiency and robustness. Key direction involves exploring hybrid training to utilize PG-TAF's staged predictions as supervision for end-to-end models.

Finally, the research should broaden its real-world impact. This means adopting the DeepSTG framework as a general diagnostic tool for other video understanding benchmarks (like action localization) and actively bridging STVG to Embodied AI and Robotics for manipulation tasks. Practical deployment requires addressing major limitations: drastically improving computational efficiency (aiming for sub-second latency) and developing strategies for handling domain shift and mitigating ethical bias in deployment scenarios.

\section{Code and Data Availability}
To maximize research impact and facilitate reproducibility, we commit to releasing all code, data, and models under permissive licenses upon the paper's acceptance. 
A small sample of OmniGround (Appendix/Sample\_OmniGround), the full evaluation code (Appendix/Code-DeepSTG), and the PG-TAF code (Appendix/Code-PGTAF) are additionally provided in the supplementary materials ZIP file.

\section{Ethical Considerations}
Our work follows the established ethical standards for dataset construction.
Our videos are collected from public platforms under permissive licenses, with no personally identifiable information included. 
Manual filtering removes inappropriate material during collection. 
The human  subjects featured in these videos appear in public contexts where the expectation of recording is diminished, consistent with the terms accepted during upload of original content. 
We annotate all synthetic samples in our benchmark to enable downstream detection and appropriate usage.

{
    \small
    \bibliographystyle{ieeenat_fullname}
    \bibliography{main}

\begin{thebibliography}{50}
\providecommand{\natexlab}[1]{#1}
\providecommand{\url}[1]{\texttt{#1}}
\expandafter\ifx\csname urlstyle\endcsname\relax
  \providecommand{\doi}[1]{doi: #1}\else
  \providecommand{\doi}{doi: \begingroup \urlstyle{rm}\Url}\fi

\bibitem[Achiam et~al.(2023)Achiam, Adler, Agarwal, Ahmad, Akkaya, Aleman, Almeida, Altenschmidt, Altman, Anadkat, et~al.]{achiam2023gpt}
Josh Achiam, Steven Adler, Sandhini Agarwal, Lama Ahmad, Ilge Akkaya, Florencia~Leoni Aleman, Diogo Almeida, Janko Altenschmidt, Sam Altman, Shyamal Anadkat, et~al.
\newblock Gpt-4 technical report.
\newblock \emph{arXiv preprint arXiv:2303.08774}, 2023.

\bibitem[Ahmad et~al.(2025)Ahmad, Heakl, Gani, Shaker, Shen, Khan, and Khan]{ahmad2025videomolmo}
Ghazi~Shazan Ahmad, Ahmed Heakl, Hanan Gani, Abdelrahman Shaker, Zhiqiang Shen, Fahad~Shahbaz Khan, and Salman Khan.
\newblock Videomolmo: Spatio-temporal grounding meets pointing.
\newblock \emph{arXiv preprint arXiv:2506.05336}, 2025.

\bibitem[Bai et~al.(2025)Bai, Chen, Liu, Wang, Ge, Song, Dang, Wang, Wang, Tang, et~al.]{bai2025qwen2}
Shuai Bai, Keqin Chen, Xuejing Liu, Jialin Wang, Wenbin Ge, Sibo Song, Kai Dang, Peng Wang, Shijie Wang, Jun Tang, et~al.
\newblock Qwen2. 5-vl technical report.
\newblock \emph{arXiv preprint arXiv:2502.13923}, 2025.

\bibitem[Burmania et~al.(2015)Burmania, Parthasarathy, and Busso]{burmania2015increasing}
Alec Burmania, Srinivas Parthasarathy, and Carlos Busso.
\newblock Increasing the reliability of crowdsourcing evaluations using online quality assessment.
\newblock \emph{IEEE Transactions on Affective Computing}, 7\penalty0 (4):\penalty0 374--388, 2015.

\bibitem[Caelles et~al.(2019)Caelles, Pont-Tuset, Perazzi, Montes, Maninis, and {Van Gool}]{Caelles_arXiv_2019}
Sergi Caelles, Jordi Pont-Tuset, Federico Perazzi, Alberto Montes, Kevis-Kokitsi Maninis, and Luc {Van Gool}.
\newblock The 2019 davis challenge on vos: Unsupervised multi-object segmentation.
\newblock \emph{arXiv:1905.00737}, 2019.

\bibitem[Cai et~al.(2025)Cai, Wang, Sun, Wang, Gu, Yin, Lin, Yang, Wei, Shi, et~al.]{cai2025has}
Zhongang Cai, Yubo Wang, Qingping Sun, Ruisi Wang, Chenyang Gu, Wanqi Yin, Zhiqian Lin, Zhitao Yang, Chen Wei, Xuanke Shi, et~al.
\newblock Has gpt-5 achieved spatial intelligence? an empirical study.
\newblock \emph{arXiv preprint arXiv:2508.13142}, 2025.

\bibitem[Cheng et~al.(2024)Cheng, Oh, Price, Lee, and Schwing]{cheng2024putting}
Ho~Kei Cheng, Seoung~Wug Oh, Brian Price, Joon-Young Lee, and Alexander Schwing.
\newblock Putting the object back into video object segmentation.
\newblock In \emph{Proceedings of the IEEE/CVF Conference on Computer Vision and Pattern Recognition}, pages 3151--3161, 2024.

\bibitem[Comanici et~al.(2025)Comanici, Bieber, Schaekermann, Pasupat, Sachdeva, Dhillon, Blistein, Ram, Zhang, Rosen, et~al.]{comanici2025gemini}
Gheorghe Comanici, Eric Bieber, Mike Schaekermann, Ice Pasupat, Noveen Sachdeva, Inderjit Dhillon, Marcel Blistein, Ori Ram, Dan Zhang, Evan Rosen, et~al.
\newblock Gemini 2.5: Pushing the frontier with advanced reasoning, multimodality, long context, and next generation agentic capabilities.
\newblock \emph{arXiv preprint arXiv:2507.06261}, 2025.

\bibitem[Ding et~al.(2023)Ding, Liu, He, Jiang, and Loy]{ding2023mevis}
Henghui Ding, Chang Liu, Shuting He, Xudong Jiang, and Chen~Change Loy.
\newblock Mevis: A large-scale benchmark for video segmentation with motion expressions.
\newblock In \emph{Proceedings of the IEEE/CVF international conference on computer vision}, pages 2694--2703, 2023.

\bibitem[Dubey et~al.(2024)Dubey, Jauhri, Pandey, Kadian, Al-Dahle, Letman, Mathur, Schelten, Yang, Fan, et~al.]{dubey2024llama}
Abhimanyu Dubey, Abhinav Jauhri, Abhinav Pandey, Abhishek Kadian, Ahmad Al-Dahle, Aiesha Letman, Akhil Mathur, Alan Schelten, Amy Yang, Angela Fan, et~al.
\newblock The llama 3 herd of models.
\newblock \emph{arXiv e-prints}, pages arXiv--2407, 2024.

\bibitem[Grauman et~al.(2022)Grauman, Westbury, Byrne, Chavis, Furnari, Girdhar, Hamburger, Jiang, Liu, Liu, et~al.]{grauman2022ego4d}
Kristen Grauman, Andrew Westbury, Eugene Byrne, Zachary Chavis, Antonino Furnari, Rohit Girdhar, Jackson Hamburger, Hao Jiang, Miao Liu, Xingyu Liu, et~al.
\newblock Ego4d: Around the world in 3,000 hours of egocentric video.
\newblock In \emph{Proceedings of the IEEE/CVF conference on computer vision and pattern recognition}, pages 18995--19012, 2022.

\bibitem[Gu et~al.(2024)Gu, Fan, Huang, Luo, and Zhang]{gu2024context}
Xin Gu, Heng Fan, Yan Huang, Tiejian Luo, and Libo Zhang.
\newblock Context-guided spatio-temporal video grounding.
\newblock In \emph{Proceedings of the IEEE/CVF Conference on Computer Vision and Pattern Recognition}, pages 18330--18339, 2024.

\bibitem[Gu et~al.(2025)Gu, Shen, Luo, Luo, Huang, Lin, Fan, and Zhang]{gu2025knowing}
Xin Gu, Yaojie Shen, Chenxi Luo, Tiejian Luo, Yan Huang, Yuewei Lin, Heng Fan, and Libo Zhang.
\newblock Knowing your target: Target-aware transformer makes better spatio-temporal video grounding.
\newblock \emph{arXiv preprint arXiv:2502.11168}, 2025.

\bibitem[Guo et~al.(2024)Guo, Liu, Li, Liu, Chen, and Tang]{guo2024trace}
Yongxin Guo, Jingyu Liu, Mingda Li, Qingbin Liu, Xi Chen, and Xiaoying Tang.
\newblock Trace: Temporal grounding video llm via causal event modeling.
\newblock \emph{arXiv preprint arXiv:2410.05643}, 2024.

\bibitem[Huang et~al.(2024)Huang, Wang, Chen, Song, and Zhu]{huang2024vtimellm}
Bin Huang, Xin Wang, Hong Chen, Zihan Song, and Wenwu Zhu.
\newblock Vtimellm: Empower llm to grasp video moments.
\newblock In \emph{Proceedings of the IEEE/CVF Conference on Computer Vision and Pattern Recognition}, pages 14271--14280, 2024.

\bibitem[Jin et~al.(2022)Jin, Yuan, Mu, et~al.]{jin2022embracing}
Yang Jin, Zehuan Yuan, Yadong Mu, et~al.
\newblock Embracing consistency: A one-stage approach for spatio-temporal video grounding.
\newblock \emph{Advances in Neural Information Processing Systems}, 35:\penalty0 29192--29204, 2022.

\bibitem[Khanam and Hussain(2024)]{khanam2024yolov11}
Rahima Khanam and Muhammad Hussain.
\newblock Yolov11: An overview of the key architectural enhancements.
\newblock \emph{arXiv preprint arXiv:2410.17725}, 2024.

\bibitem[Li et~al.(2025)Li, Chen, Wei, Huang, Hui, Gao, Wei, and Liu]{li2025llava}
Hongyu Li, Jinyu Chen, Ziyu Wei, Shaofei Huang, Tianrui Hui, Jialin Gao, Xiaoming Wei, and Si Liu.
\newblock Llava-st: A multimodal large language model for fine-grained spatial-temporal understanding.
\newblock In \emph{Proceedings of the Computer Vision and Pattern Recognition Conference}, pages 8592--8603, 2025.

\bibitem[Li et~al.(2024)Li, Xu, Zhang, Song, Cai, Qi, Zhou, Pan, Li, Vu, et~al.]{li2024groundinggpt}
Zhaowei Li, Qi Xu, Dong Zhang, Hang Song, Yiqing Cai, Qi Qi, Ran Zhou, Junting Pan, Zefeng Li, Van~Tu Vu, et~al.
\newblock Groundinggpt: Language enhanced multi-modal grounding model.
\newblock \emph{arXiv preprint arXiv:2401.06071}, 2024.

\bibitem[Liang et~al.(2025)Liang, Zhong, Hu, Tao, and Wang]{liang2025fine}
Shuo Liang, Yiwu Zhong, Zi-Yuan Hu, Yeyao Tao, and Liwei Wang.
\newblock Fine-grained spatiotemporal grounding on egocentric videos.
\newblock In \emph{Proceedings of the IEEE/CVF International Conference on Computer Vision}, pages 9385--9395, 2025.

\bibitem[Lin(2002)]{lin2002divergence}
Jianhua Lin.
\newblock Divergence measures based on the shannon entropy.
\newblock \emph{IEEE Transactions on Information theory}, 37\penalty0 (1):\penalty0 145--151, 2002.

\bibitem[Lin et~al.(2014)Lin, Maire, Belongie, Hays, Perona, Ramanan, Doll{\'a}r, and Zitnick]{lin2014microsoft}
Tsung-Yi Lin, Michael Maire, Serge Belongie, James Hays, Pietro Perona, Deva Ramanan, Piotr Doll{\'a}r, and C~Lawrence Zitnick.
\newblock Microsoft coco: Common objects in context.
\newblock In \emph{European conference on computer vision}, pages 740--755. Springer, 2014.

\bibitem[Liu et~al.(2024)Liu, Feng, Xue, Wang, Wu, Lu, Zhao, Deng, Zhang, Ruan, et~al.]{liu2024deepseek}
Aixin Liu, Bei Feng, Bing Xue, Bingxuan Wang, Bochao Wu, Chengda Lu, Chenggang Zhao, Chengqi Deng, Chenyu Zhang, Chong Ruan, et~al.
\newblock Deepseek-v3 technical report.
\newblock \emph{arXiv preprint arXiv:2412.19437}, 2024.

\bibitem[Lotfian and Busso(2017)]{lotfian2017building}
Reza Lotfian and Carlos Busso.
\newblock Building naturalistic emotionally balanced speech corpus by retrieving emotional speech from existing podcast recordings.
\newblock \emph{IEEE Transactions on Affective Computing}, 10\penalty0 (4):\penalty0 471--483, 2017.

\bibitem[Ma et~al.(2024)Ma, Jiang, Wu, Yuan, and Qi]{ma2024groma}
Chuofan Ma, Yi Jiang, Jiannan Wu, Zehuan Yuan, and Xiaojuan Qi.
\newblock Groma: Localized visual tokenization for grounding multimodal large language models.
\newblock In \emph{European Conference on Computer Vision}, pages 417--435. Springer, 2024.

\bibitem[Peng et~al.(2024)Peng, Gao, Liu, Li, Dong, Zhang, Fan, and Zhang]{peng2024vasttrack}
Liang Peng, Junyuan Gao, Xinran Liu, Weihong Li, Shaohua Dong, Zhipeng Zhang, Heng Fan, and Libo Zhang.
\newblock Vasttrack: Vast category visual object tracking.
\newblock \emph{Advances in Neural Information Processing Systems}, 37:\penalty0 130797--130818, 2024.

\bibitem[Seo et~al.(2020)Seo, Lee, and Han]{seo2020urvos}
Seonguk Seo, Joon-Young Lee, and Bohyung Han.
\newblock Urvos: Unified referring video object segmentation network with a large-scale benchmark.
\newblock In \emph{European conference on computer vision}, pages 208--223. Springer, 2020.

\bibitem[Shang et~al.(2019)Shang, Di, Xiao, Cao, Yang, and Chua]{shang2019annotating}
Xindi Shang, Donglin Di, Junbin Xiao, Yu Cao, Xun Yang, and Tat-Seng Chua.
\newblock Annotating objects and relations in user-generated videos.
\newblock In \emph{Proceedings of the 2019 on International Conference on Multimedia Retrieval}, pages 279--287, 2019.

\bibitem[Shao et~al.(2019)Shao, Li, Zhang, Peng, Yu, Zhang, Li, and Sun]{shao2019objects365}
Shuai Shao, Zeming Li, Tianyuan Zhang, Chao Peng, Gang Yu, Xiangyu Zhang, Jing Li, and Jian Sun.
\newblock Objects365: A large-scale, high-quality dataset for object detection.
\newblock In \emph{Proceedings of the IEEE/CVF international conference on computer vision}, pages 8430--8439, 2019.

\bibitem[Simon(2001)]{simon2001kalman}
Dan Simon.
\newblock Kalman filtering.
\newblock \emph{Embedded systems programming}, 14\penalty0 (6):\penalty0 72--79, 2001.

\bibitem[Su et~al.(2021)Su, Yu, and Xu]{su2021stvgbert}
Rui Su, Qian Yu, and Dong Xu.
\newblock Stvgbert: A visual-linguistic transformer based framework for spatio-temporal video grounding.
\newblock In \emph{Proceedings of the IEEE/CVF International Conference on Computer Vision}, pages 1533--1542, 2021.

\bibitem[Sun et~al.(2024)Sun, Fang, Wu, Zhang, Zang, Kong, Xiong, Lin, and Wang]{sun2024alpha}
Zeyi Sun, Ye Fang, Tong Wu, Pan Zhang, Yuhang Zang, Shu Kong, Yuanjun Xiong, Dahua Lin, and Jiaqi Wang.
\newblock Alpha-clip: A clip model focusing on wherever you want.
\newblock In \emph{Proceedings of the IEEE/CVF conference on computer vision and pattern recognition}, pages 13019--13029, 2024.

\bibitem[Tang et~al.(2021)Tang, Liao, Liu, Li, Jin, Jiang, Yu, and Xu]{tang2021human}
Zongheng Tang, Yue Liao, Si Liu, Guanbin Li, Xiaojie Jin, Hongxu Jiang, Qian Yu, and Dong Xu.
\newblock Human-centric spatio-temporal video grounding with visual transformers.
\newblock \emph{IEEE Transactions on Circuits and Systems for Video Technology}, 32\penalty0 (12):\penalty0 8238--8249, 2021.

\bibitem[Team et~al.(2025)Team, Yang, Wen, Liu, Chu, Song, Rao, Yi, Li, Zang, et~al.]{team2025kwai}
Kwai~Keye Team, Biao Yang, Bin Wen, Changyi Liu, Chenglong Chu, Chengru Song, Chongling Rao, Chuan Yi, Da Li, Dunju Zang, et~al.
\newblock Kwai keye-vl technical report.
\newblock \emph{arXiv preprint arXiv:2507.01949}, 2025.

\bibitem[Vasiliev(2020)]{vasiliev2020natural}
Yuli Vasiliev.
\newblock \emph{Natural language processing with Python and spaCy: A practical introduction}.
\newblock No Starch Press, 2020.

\bibitem[Videnovic et~al.(2025)Videnovic, Lukezic, and Kristan]{videnovic2025distractor}
Jovana Videnovic, Alan Lukezic, and Matej Kristan.
\newblock A distractor-aware memory for visual object tracking with sam2.
\newblock In \emph{Proceedings of the Computer Vision and Pattern Recognition Conference}, pages 24255--24264, 2025.

\bibitem[Vishal et~al.(2024)Vishal, Basina, Choudhary, and Chakravarthi]{vishal2024eyes}
Joseph~Raj Vishal, Divesh Basina, Aarya Choudhary, and Bharatesh Chakravarthi.
\newblock Eyes on the road: State-of-the-art video question answering models assessment for traffic monitoring tasks.
\newblock \emph{arXiv preprint arXiv:2412.01132}, 2024.

\bibitem[Wan et~al.(2025)Wan, Wang, Ai, Wen, Mao, Xie, Chen, Yu, Zhao, Yang, et~al.]{wan2025wan}
Team Wan, Ang Wang, Baole Ai, Bin Wen, Chaojie Mao, Chen-Wei Xie, Di Chen, Feiwu Yu, Haiming Zhao, Jianxiao Yang, et~al.
\newblock Wan: Open and advanced large-scale video generative models.
\newblock \emph{arXiv preprint arXiv:2503.20314}, 2025.

\bibitem[Wang et~al.(2025{\natexlab{a}})Wang, Zhang, Liu, Li, Ge, Xie, and Zhang]{wang2025spacevllm}
Jiankang Wang, Zhihan Zhang, Zhihang Liu, Yang Li, Jiannan Ge, Hongtao Xie, and Yongdong Zhang.
\newblock Spacevllm: Endowing multimodal large language model with spatio-temporal video grounding capability.
\newblock \emph{arXiv preprint arXiv:2503.13983}, 2025{\natexlab{a}}.

\bibitem[Wang et~al.(2025{\natexlab{b}})Wang, Hu, Li, Safari, and Yang]{wang2025capabilities}
Shansong Wang, Mingzhe Hu, Qiang Li, Mojtaba Safari, and Xiaofeng Yang.
\newblock Capabilities of gpt-5 on multimodal medical reasoning.
\newblock \emph{arXiv preprint arXiv:2508.08224}, 2025{\natexlab{b}}.

\bibitem[Wang et~al.(2023{\natexlab{a}})Wang, Chen, Chen, Wu, Zhu, Zeng, Luo, Lu, Zhou, Qiao, et~al.]{wang2023visionllm}
Wenhai Wang, Zhe Chen, Xiaokang Chen, Jiannan Wu, Xizhou Zhu, Gang Zeng, Ping Luo, Tong Lu, Jie Zhou, Yu Qiao, et~al.
\newblock Visionllm: Large language model is also an open-ended decoder for vision-centric tasks.
\newblock \emph{Advances in Neural Information Processing Systems}, 36:\penalty0 61501--61513, 2023{\natexlab{a}}.

\bibitem[Wang et~al.(2023{\natexlab{b}})Wang, Liu, Su, and Nie]{WANG2023efficient}
Weikang Wang, Jing Liu, Yuting Su, and Weizhi Nie.
\newblock Efficient spatio-temporal video grounding with semantic-guided feature decomposition.
\newblock In \emph{Proceedings of the 31st ACM International Conference on Multimedia}, page 4867–4876, New York, NY, USA, 2023{\natexlab{b}}. Association for Computing Machinery.

\bibitem[Wu(2012)]{wu2012maximum}
Nailong Wu.
\newblock \emph{The maximum entropy method}.
\newblock Springer Science \& Business Media, 2012.

\bibitem[Yang et~al.(2022)Yang, Miech, Sivic, Laptev, and Schmid]{yang2022tubedetr}
Antoine Yang, Antoine Miech, Josef Sivic, Ivan Laptev, and Cordelia Schmid.
\newblock Tubedetr: Spatio-temporal video grounding with transformers.
\newblock In \emph{Proceedings of the IEEE/CVF Conference on Computer Vision and Pattern Recognition}, pages 16442--16453, 2022.

\bibitem[Yang et~al.(2025{\natexlab{a}})Yang, Li, Yang, Zhang, Hui, Zheng, Yu, Gao, Huang, Lv, et~al.]{yang2025qwen3}
An Yang, Anfeng Li, Baosong Yang, Beichen Zhang, Binyuan Hui, Bo Zheng, Bowen Yu, Chang Gao, Chengen Huang, Chenxu Lv, et~al.
\newblock Qwen3 technical report.
\newblock \emph{arXiv preprint arXiv:2505.09388}, 2025{\natexlab{a}}.

\bibitem[Yang et~al.(2025{\natexlab{b}})Yang, Xu, Kaiser, Cheng, Rosenhahn, and Yang]{yang2025multio}
Yi Yang, Yiming Xu, Timo Kaiser, Hao Cheng, Bodo Rosenhahn, and Michael~Ying Yang.
\newblock Multi-object tracking retrieval with llava-video: A training-free solution to mot25-stag challenge, 2025{\natexlab{b}}.

\bibitem[Zhang et~al.(2025)Zhang, Li, Zhang, Pu, Cahyono, Hu, Liu, Zhang, Yang, Li, et~al.]{zhang2025lmms}
Kaichen Zhang, Bo Li, Peiyuan Zhang, Fanyi Pu, Joshua~Adrian Cahyono, Kairui Hu, Shuai Liu, Yuanhan Zhang, Jingkang Yang, Chunyuan Li, et~al.
\newblock Lmms-eval: Reality check on the evaluation of large multimodal models.
\newblock In \emph{Findings of the Association for Computational Linguistics: NAACL 2025}, pages 881--916, 2025.

\bibitem[Zhang et~al.(2024)Zhang, Cheng, Zhu, Hu, Liu, Liu, Ran, Chen, Liu, and Wang]{zhang2024evf}
Yuxuan Zhang, Tianheng Cheng, Lianghui Zhu, Rui Hu, Lei Liu, Heng Liu, Longjin Ran, Xiaoxin Chen, Wenyu Liu, and Xinggang Wang.
\newblock Evf-sam: Early vision-language fusion for text-prompted segment anything model.
\newblock \emph{arXiv preprint arXiv:2406.20076}, 2024.

\bibitem[Zhang et~al.(2020)Zhang, Zhao, Zhao, Wang, Liu, and Gao]{zhang2020does}
Zhu Zhang, Zhou Zhao, Yang Zhao, Qi Wang, Huasheng Liu, and Lianli Gao.
\newblock Where does it exist: Spatio-temporal video grounding for multi-form sentences.
\newblock In \emph{Proceedings of the IEEE/CVF Conference on Computer Vision and Pattern Recognition}, pages 10668--10677, 2020.

\bibitem[Zhou et~al.(2023)Zhou, Stefanidis, Jiang, Sui, and Feng]{zhou2023language}
Mian Zhou, Angelos Stefanidis, Nan Jiang, Zezhou Sui, and Zhikun Feng.
\newblock Language-led visual grounding for human computer interaction.
\newblock \emph{Preprints}, 2023.

\end{thebibliography}
}


\end{document}